\pdfoutput=1

\documentclass[11pt]{article}

\usepackage[final]{acl}

\usepackage{times}
\usepackage{latexsym}

\usepackage[T1]{fontenc}

\usepackage[utf8]{inputenc}

\usepackage{microtype}

\usepackage{inconsolata}

\usepackage{graphicx}

\usepackage{soul}
\usepackage{multirow}
\usepackage{booktabs}

\usepackage{amsmath, amssymb}

\usepackage{xcolor}
\usepackage{colortbl}

\usepackage{subcaption}

\usepackage{booktabs}
\usepackage[utf8]{inputenc}
\usepackage{longtable}

\newcommand{\df}{\textcolor{black}}

\newcommand{\ga}{\textcolor{black}}

\newcommand{\repo}{
\url{https://github.com/hlt-mt/speech-translation-gender}
}

\title{Different Speech Translation Models \\Encode and Translate Speaker Gender Differently}

\author{
Dennis Fucci$^\alpha$$^\beta$, Marco Gaido$^\beta$, Matteo Negri$^\beta$, \\
\textbf{Luisa Bentivogli}$^\beta$, \textbf{André F. T. Martins}$^{\gamma}$$^{\delta}$$^{\epsilon}$, \textbf{Giuseppe Attanasio}$^\epsilon$ \\ $^\alpha$University of Trento, Italy
 $^\beta$Fondazione Bruno Kessler, Italy  \\
 $^\gamma$Unbabel, Portugal
$^\delta$Instituto Superior Técnico, Portugal \\
$^\epsilon$Instituto de Telecomunicações, Portugal \\
   \texttt{dfucci@fbk.eu}
   }

\begin{document}
\maketitle
\begin{abstract}
Recent studies on interpreting the hidden states of speech models have shown their ability to capture speaker-specific features, including gender. Does this finding also hold for speech translation (ST) models? If so, what are the implications for the speaker's gender assignment in translation? We address these questions from an interpretability perspective, using probing methods to assess gender encoding across diverse ST models. Results on three language directions (English $\rightarrow$ French/Italian/Spanish) indicate that while traditional encoder-decoder models capture gender information, newer architectures---integrating a speech encoder with a machine translation system via adapters---do not. We also demonstrate that low gender encoding capabilities result in systems' tendency toward a masculine default, a translation bias that is more pronounced in newer architectures.\footnote{
Code available under Apache 2.0 at \repo.}
\end{abstract}

\section{Introduction}

Recent research in speech representation learning shows that models capture phonetic and speaker-related features in their internal representations \citep[e.g.,][]{prasad-jyothi-2020-accents, cormac-english-etal-2022-domain, pasad_etal_2023, yang23v_interspeech, choi24b_interspeech, shen-etal-2024-encoding, chowdhury-2024, waheed2024speechfoundationmodelslearn}. 
Building on sociophonetic studies that show how sociocultural and physiological factors shape voice production and perception \citep[e.g.,][]{coleman-1976-voice, fuchs2010differences, azul2013voices, leung2018voice}, recent research has explored how speaker \textit{gender} is encoded in speech models. Studies have found evidence of gender encoding in self-supervised models \citep{chowdhury-2024, deseyssel22_interspeech, guillaume24_interspeech} as well as in automatic speech recognition systems \citep{krishnan24_interspeech, attanasio-etal-2024-twists}.
Despite growing interest, an analysis of gender encoding and its impact in speech translation (ST) remains largely absent.

Gender plays a crucial role in ST, particularly when translating from 
\textit{notional}  to \textit{grammatical}
gender languages. Here, models must infer gender from context and correctly apply inflections. Research suggests that encoder-decoder ST systems may use acoustic cues to assign grammatical gender to words referring to the speaker (e.g., En: ``I was \underline{born} in...'' $\rightarrow$ Fr: ``Je suis \underline{né/née} à...''), yet masculine defaults remain common \citep{bentivogli-etal-2020-gender, gaido-etal-2020-breeding}.\footnote{\df{While inferring gender from acoustic cues is common---even among humans---it should not be treated as the default. This paper examines whether models exhibit similar behavior (for further discussion, see §\ref{sec:ethics_statement}).}}
Moreover, as the ST field evolves, traditional encoder-decoder systems are being supplemented by newer architectures that integrate pretrained speech encoders into machine translation 
models 
via adapters
(speech+MT).
This raises key questions: \textbf{\textit{whether} gender information is encoded and used by current ST systems}, and \textbf{\textit{how} architectural variations influence this process}. 
Addressing these questions can shed light on gender-biased behaviors, 
such as the systematic preference for masculine defaults.

We investigate these aspects using probing, an established interpretability method \df{\citep{conneau-etal-2018-cram, probing_belinkov}.} We train probes to predict the speaker's gender\footnote{We limit our analyses to binary gender, though we recognize that gender exists on a continuum (see \S \ref{sec:ethics_statement}).}
from hidden states in traditional encoder-decoder and speech+MT models. Then, by analyzing translations from English into Spanish, French, and Italian, we assess how gender encoding influences the translation of gender-marked terms referring to the speaker.
Our results demonstrate that encoder-decoder models encode gender information, whereas speech+MT models encode it minimally, or do not encode it at all. 
Moreover, the ability to extract gender from hidden states correlates with how accurately models inflect gender in speaker-referred words.
Strong probing performance is a proxy for high gender translation accuracy, while weak probing performance correlates with translations defaulting to masculine.
\df{These findings suggest that ST models do not uniformly rely on acoustic cues to translate speaker-referred expressions.}

Our findings are compelling for several reasons.
On the one hand, they show that encoding gender yields fairer systems for female users, challenging the conventional understanding in NLP that scrubbing gender information leads to fairer outcomes \cite{sun-etal-2019-mitigating}.
On the other hand, relying on biometric markers for automatic decision-making contrasts broader and established ethical principles. We discuss the matter in \S \ref{sec:ethics_statement}.

\begin{table*}[t!]
\centering
\footnotesize
\setlength{\tabcolsep}{4.5pt}
\begin{tabular}{ll|c|ccc|ccc|ccc|c}
\toprule
 & & \texttt{\textbf{test-generic}} & \multicolumn{10}{c}{\texttt{\textbf{test-speaker}}} \\
 & &  & \multicolumn{3}{c|}{\textbf{en$\rightarrow$es}} & \multicolumn{3}{c|}{\textbf{en$\rightarrow$fr}} & \multicolumn{3}{c|}{\textbf{en$\rightarrow$it}} & \textbf{Avg.} \\
 & & \textbf{\texttt{All}} & \textbf{\texttt{She}} & \textbf{\texttt{He}} & \textbf{\texttt{All}} & \textbf{\texttt{She}} & \textbf{\texttt{He}} & \textbf{\texttt{All}} & \textbf{\texttt{She}} & \textbf{\texttt{He}} & \textbf{\texttt{All}} & \textbf{\texttt{All}} \\
\midrule
\multirow{2}{*}{\texttt{Seamless}} & \texttt{post-ad} & \cellcolor{gray!20}59.76 & \cellcolor{gray!20}90.41 & \cellcolor{gray!20}24.30 & \cellcolor{gray!20}51.72 & \cellcolor{gray!20}86.18 & \cellcolor{gray!20}29.55 & \cellcolor{gray!20}54.51 & \cellcolor{gray!20}85.98 & \cellcolor{gray!20}32.14 & \cellcolor{gray!20}55.63 & \cellcolor{gray!20}53.95 \\
 & \texttt{pre-ad} & 75.53 & 85.61 & 51.41 & 67.32 & 81.25 & 54.64 & 67.52 & 83.76 & 58.57 & 70.58 & 68.47 \\
\midrule
\multirow{2}{*}{\texttt{ZeroSwot}} & \texttt{post-ad} & \cellcolor{gray!20}59.57 & \cellcolor{gray!20}85.98 & \cellcolor{gray!20}38.73 & \cellcolor{gray!20}61.80 & \cellcolor{gray!20}90.13 & \cellcolor{gray!20}37.80 & \cellcolor{gray!20}61.62 & \cellcolor{gray!20}86.35 & \cellcolor{gray!20}39.64 & \cellcolor{gray!20}60.65 & \cellcolor{gray!20}61.36 \\
 & \texttt{pre-ad} & 89.60 & 96.68 & 84.15 & 90.25 & 95.72 & 84.19 & 90.02 & 92.99 & 84.29 & 88.56 & 89.61 \\
\midrule
\multicolumn{2}{c|}{\texttt{enc-dec}} & \cellcolor{gray!20}92.21 & \cellcolor{gray!20}99.26 & \cellcolor{gray!20}87.32 & \cellcolor{gray!20}93.14 & \cellcolor{gray!20}99.67 & \cellcolor{gray!20}89.35 & \cellcolor{gray!20}94.59 & \cellcolor{gray!20}98.89 & \cellcolor{gray!20}93.57 & \cellcolor{gray!20}96.19 & \cellcolor{gray!20}94.64 \\
\bottomrule
\end{tabular}
\caption{
Scores for gender probing (macro F1 for \texttt{All}, recall for \texttt{She}/\texttt{He}) \df{across all configurations.} 
}
\label{tab:results_probing}
\end{table*}

\section{Gender Probing}
\label{sec:gender_probing}

The encoder of a speech model maps an arbitrary audio signal into a sequence of \textit{d}-dimensional hidden states \( \mathbf{X} = \langle \mathbf{x}_1, \mathbf{x}_2, \dots, \mathbf{x}_L \rangle \), with \( \mathbf{x}_l \in \mathbb{R}^{d} \) being the model’s speech representation at position \( l \) and \( L \) the output sequence length.
To evaluate whether and to which extent \( \mathbf{X} \) encodes a specific attribute, classification \textit{probes} \cite{probing_belinkov} can be trained on states in \( \mathbf{X} \), with classification accuracy indicating how well the attribute is encoded.

To obtain a single classification label for the whole sequence, previous works on gender encoding reduce its length \( L \) using mean or max pooling before training logistic probes \cite{deseyssel22_interspeech, chowdhury-2024, guillaume24_interspeech, krishnan24_interspeech} or non-linear probes \cite{krishnan24_interspeech}.
However, pooling may obscure positional variations in gender encoding, potentially weakening classification performance.
Alternative approaches train separate probes on individual states \( \mathbf{x}_l \) at specific positions \( l \), assuming gender is not uniformly encoded across \( L \).
While more fine-grained than pooling, these methods have only considered a fixed, small set of relative positions \cite{krishnan24_interspeech} or do not support sequences of arbitrary length \citep{attanasio-etal-2024-twists}.

To avoid the limitations above, we design an attention-based probe.
We draw inspiration from the Q-Former
\cite{qformer},
which maps a sequence of variable length into a fixed number of vectors by means of an attention mechanism where the query is a fixed sequence of learnable vectors.
Similarly, our probe 
leverages a single learnable query
to selectively extract gender information from hidden states.
The sequence of hidden states \( \mathbf{X} \) is projected into key (\( \mathbf{K} \in \mathbb{R}^{L \times d} \)) and value (\( \mathbf{V} \in \mathbb{R}^{L \times d} \)) matrices using learnable weight matrices \( \mathbf{W}_K \in \mathbb{R}^{d \times d} \) and \( \mathbf{W}_V \in \mathbb{R}^{d \times d} \), respectively. We then compute scaled-dot attention using \( \mathbf{K} \), \( \mathbf{V} \), and a learnable query vector 
\( \mathbf{q} \in \mathbb{R}^{d} \). 
The output representation \( \mathbf{o} \in \mathbb{R}^{d} \) is passed through a linear layer to compute class logits, and final probabilities are obtained via softmax.\footnote{See Appendix \ref{app:setting} for implementation details.}
The attention weights \( \mathbf{a} \in \mathbb{R}^{L} \) instead indicate, for each input sequence, which hidden states contribute most to gender encoding.
\df{By using attention to selectively extract gender information from the entire input, followed by a single linear classification layer, the architecture remains simple---an important property for avoiding spurious correlations, as recommended for probing classifiers \cite{probing_belinkov}. At the same time, this attention-based design is more expressive than basic linear models \cite{hewitt-liang-2019-designing},
and serves as a proxy for how ST decoders access encoder states---namely, through the cross-attention mechanism.}
For a comparison with probes from previous works, see Appendix \ref{app:probe_comparison}.

\section{Experimental Setup}
\label{sec:exp_setup}

\paragraph{Models} We evaluate one
ST 
encoder-decoder and two speech+MT models 
(full details in Appendix \ref{app:setting})
for 
en$\rightarrow$es/fr/it translation.
As encoder-decoder model (\texttt{enc-dec}), we use a Transformer-based model by \citet{wang2020fairseqs2t},
trained on ST data from MuST-C \cite{mustc}. The speech+MT models are SeamlessM4T (\texttt{Seamless}---\citealp{seamless}) and ZeroSwot (\texttt{ZeroSwot}---\citealp{tsiamas-etal-2024-pushing}).
We probe the model states at different locations.
In \texttt{enc-dec}, we use the encoder outputs.
In speech+MT models, to assess the impact of adapters---which compress and map the speech encoder's outputs to the MT model's embedding space---we probe hidden states before (\texttt{pre-ad}) and after (\texttt{post-ad}) the adapters.

\paragraph{Data} To train and evaluate probes, we use the MuST-C corpus, \df{which includes speaker gender annotations based on self-declared pronouns (\texttt{He}/\texttt{She}) at the time of data collection \citep{gaido-etal-2020-breeding}, thereby avoiding the risk of misgendering (see \S \ref{sec:ethics_statement}).}
For training, we randomly sample audio segments from the MuST-C training set (en$\rightarrow$es section) to create 
gender-balanced training (\texttt{train}) and validation (\texttt{dev}) sets. 
For testing, we use two datasets.\footnote{No speaker overlaps exist across \texttt{train}, \texttt{dev}, and test sets. For details on data statistics, see Appendix \ref{sec:data_statistics}.}
\texttt{Test-generic} consists of generic 
utterances taken from the validation and test sets of MuST-C. Since these sets are imbalanced across gender classes, \texttt{test-generic} is also inherently unbalanced. 
\texttt{Test-speaker} is drawn from MuST-SHE \cite{bentivogli-etal-2020-gender}, 
\df{a corpus specifically designed to evaluate gender translation related to human referents from English into Spanish, French, and Italian.}
In MuST-SHE reference translations, each gender-marked word---corresponding to a neutral expression in English---is annotated with its opposite (wrong) gender form. We use the portion of MuST-SHE containing first-person references (e.g. En. ``I was born'' $\rightarrow$ 
\df{Fr. ``Je suis née\textsuperscript{F}<né\textsuperscript{M}>''),
\df{which is balanced across the two genders}.} 
Through \texttt{test-speaker}, we test gender encoding, gender translation, and their relationship, as the only cue for determining gender translation is the speaker's voice characteristics.

\paragraph{Evaluation} We use macro F1 for overall probing performance, and recall for individual balanced classes. 
For overall ST quality, we use COMET \cite{rei-etal-2020-comet}.\footnote{We use the \texttt{Unbabel/wmt22-comet-da} model.} For gender translation, we use the MuST-SHE evaluation script. It calculates \textit{accuracy} as the percentage of speaker-referred words generated with the correct gender\df{,} out of all generated speaker-referred words. Instances where the model fails to predict the 
\df{speaker-referred words---regardless}
of gender---are excluded 
\df{from the accuracy calculation.}
The percentage of words included in the accuracy computation is reported as \textit{coverage} \cite{gaido-etal-2020-breeding}.
\df{Although out-of-coverage instances are excluded from gender accuracy computations, they can still convey gender information---for example, when the generated term is a gendered synonym. However, through manual analysis of MuST-SHE translations by ST systems, \citet{savoldi-etal-2022-morphosyntactic} demonstrated that out-of-coverage cases are typically either gendered synonyms that follow the same accuracy patterns as in-coverage examples, or mistranslations, where gender assessment is not applicable. 
We confirmed that this pattern also holds for the ST models investigated in this study through a manual analysis of out-of-coverage outputs, detailed in Appendix~\ref{sec:manual_eval}.}

\begin{table*}[t!]
\centering
\footnotesize
\begin{tabular}{l|l|ccc|ccc|ccc}
\toprule
 & & \multicolumn{3}{c|}{\textbf{\texttt{She}}} & \multicolumn{3}{c|}{\textbf{\texttt{He}}} & \multicolumn{3}{c}{\textbf{\texttt{All}}} \\
 & & \textbf{COMET} & \textbf{Cov.} & \textbf{Acc.} & \textbf{COMET} & \textbf{Cov.} & \textbf{Acc.} & \textbf{COMET} & \textbf{Cov.} & \textbf{Acc.} \\
\midrule
\multirow{4}{*}{\texttt{Seamless}} & \textbf{en$\rightarrow$es} & 80.81 & 70.37 & \cellcolor{gray!20}12.09 & 81.69 & 70.56 & \cellcolor{gray!20}90.55 & 81.26 & 70.47 & \cellcolor{gray!20}53.62 \\
& \textbf{en$\rightarrow$fr} & 78.31 & 59.61 & \cellcolor{gray!20}17.60 & 79.50 & 59.66 & \cellcolor{gray!20}88.84 & 78.89 & 59.63 & \cellcolor{gray!20}53.29\\
& \textbf{en$\rightarrow$it} & 79.35 & 57.11 & \cellcolor{gray!20}13.30 & 82.45 & 57.52 & \cellcolor{gray!20}91.36 & 80.93 & 57.32 & \cellcolor{gray!20}53.15 \\
& avg. & 79.49 & 62.36 & \cellcolor{gray!20}14.33 & 81.21 & 42.61 & \cellcolor{gray!20}90.25 & 80.36 & 62.47 & \cellcolor{gray!20}53.35 \\
\midrule
\multirow{4}{*}{\texttt{ZeroSwot}} & \textbf{en$\rightarrow$es} & 85.57 & 78.04 & \cellcolor{gray!20}51.50 & 83.56 & 73.72 & \cellcolor{gray!20}76.73 & 84.54 & 75.79 & \cellcolor{gray!20}64.46 \\
& \textbf{en$\rightarrow$fr} & 82.99 & 65.52 & \cellcolor{gray!20}54.38 & 82.20 & 64.55 & \cellcolor{gray!20}70.11 & 82.60 & 65.03 & \cellcolor{gray!20}62.20 \\
& \textbf{en$\rightarrow$it} & 84.62 & 65.23 & \cellcolor{gray!20}46.18 & 84.71 & 60.19 & \cellcolor{gray!20}77.87 & 84.67 & 62.66 & \cellcolor{gray!20}61.75 \\
& avg. & 84.39 & 69.60 & \cellcolor{gray!20}50.69 & 83.49 & 66.15 & \cellcolor{gray!20}74.90 & 83.94 & 67.83 & \cellcolor{gray!20}62.80 \\
\midrule
\multirow{4}{*}{\texttt{enc-dec}} & \textbf{en$\rightarrow$es} & 77.03 & 66.93 & \cellcolor{gray!20}81.47 & 75.64 & 65.69 & \cellcolor{gray!20}90.94 & 76.32 & 66.29 & \cellcolor{gray!20}86.45 \\
& \textbf{en$\rightarrow$fr} & 71.77 & 53.20 & \cellcolor{gray!20}77.68 & 73.68 & 53.79 & \cellcolor{gray!20}91.34 & 72.70 & 53.50 & \cellcolor{gray!20}84.62 \\
& \textbf{en$\rightarrow$it} & 74.56 & 55.33 & \cellcolor{gray!20}76.44 & 76.02 & 54.85 & \cellcolor{gray!20}94.47 & 75.30 & 55.09 & \cellcolor{gray!20}85.65 \\
& avg. & 74.45 & 58.49 & \cellcolor{gray!20}78.53 & 75.11 & 58.11 & \cellcolor{gray!20}92.25 & 74.77 & 58.29 & \cellcolor{gray!20}85.57 \\
\bottomrule
\end{tabular}
\caption{Translation performance in quality (COMET), gender coverage, and accuracy on \texttt{test-speaker} \df{for all three language pairs, along with average scores}.}
\label{tab:performance}
\end{table*}

\section{Results}

\subsection{Gender Probing}

Table \ref{tab:results_probing} presents the macro F1 and single-class recall scores for the probes on \texttt{test-generic} and \texttt{test-speaker}.
We first compare the results of probes tested on the final speech representations of \texttt{enc-dec} and 
\texttt{Seamless/ZeroSwot post-ad}.
Overall, probes trained on the hidden states of \texttt{enc-dec} achieve high accuracy, with F1 scores peaking at $96.19$ 
\df{on \texttt{test-speaker} (\texttt{All} set, en$\rightarrow$it)
and scoring $92.21$ on \texttt{test-generic}.}
In contrast, probes trained on \texttt{Seamless/ZeroSwot post-ad}
yield significantly lower scores, not exceeding $59.76$ on \texttt{test-generic} and $61.80$ on \texttt{test-speaker} (en$\rightarrow$es), respectively. 
Focusing on \texttt{test-speaker}, we note that
\texttt{ZeroSwot post-ad} retains slightly more gender information than \texttt{Seamless post-ad}.
Single-class recall scores are higher for \texttt{She} than \texttt{He} across all probes, especially in the weakest probe trained on \texttt{Seamless post-ad}.
While this might suggest that the \texttt{He} class is harder to extract, the strong skew toward the \texttt{She} class in  \texttt{Seamless} and \texttt{ZeroSwot post-ad} likely results from the probes' overall difficulty in extracting gender from these hidden states.
Overall, these findings indicate that \textbf{speaker's gender encoding
capability is high in the speech representations of the traditional encoder-decoder model, but significantly lower in speech+MT models}. 

We now analyze the impact of the adapters in \texttt{Seamless} and \texttt{ZeroSwot}. 
Looking at \texttt{pre-ad} results, \texttt{ZeroSwot}
exhibits stronger probing performance 
compared to \texttt{Seamless} (avg. $89.61$ vs $68.47$ on \texttt{test-speaker}).
This difference may stem from their training strategies: \texttt{Seamless} is jointly trained on speech and text, while \texttt{ZeroSwot} trains only the speech encoder, 
keeping the rest of the model frozen.
Comparing \texttt{pre-ad} and \texttt{post-ad} results, we observe a substantial decrease in probing performance, with a $\sim21\%$ drop in F1 for \texttt{Seamless} and $\sim32\%$ for \texttt{ZeroSwot} after the adapters, across both test sets.
These results indicate that 
\textbf{mapping speech representation into the MT embedding space via adapters significantly removes gender information in the speech+MT models.}

In summary, gender encoding varies across models. Speech+MT systems show lower encoding capability, particularly after the adapters, while the encoder-decoder model shows higher encoding capability, similarly to speech models explored in previous works. 
Additionally, we conducted an analysis to examine how gender information is distributed across the sequence length (see Appendix~\ref{app:gender_in_seq_len}). Interestingly, and consistent with prior findings on ASR models, we observe that ST models primarily encode gender in the early positions of the sequence.

\subsection{Speaker's Gender Translation}
\label{sec:results_gender_translation}

\begin{figure}[t!]
    \centering
    \includegraphics[width=0.45\textwidth]{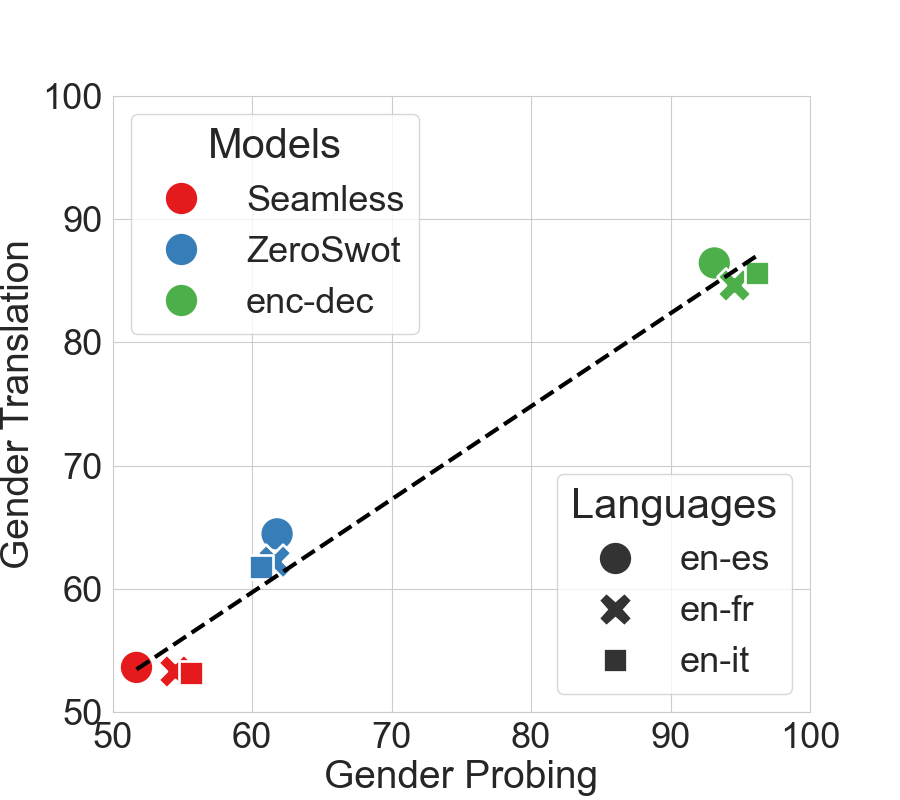}
    \caption{
    \df{Correlation between overall gender probing performance (macro F1) and gender translation accuracy across models and languages on \texttt{test-speaker}}.}
    \label{fig:correlation}
\end{figure}

Table \ref{tab:performance} presents translation scores on \texttt{test-speaker} across \df{the} three ST models and language directions.
Overall, \texttt{ZeroSwot} achieves the best translation quality, followed by \texttt{Seamless} and \texttt{enc-dec} (avg. COMET scores are respectively $83.94$, $80.36$, and $74.77$). Coverage scores align with COMET trends, indicating that higher translation quality increases the likelihood that system outputs include 
\df{the speaker-referred words annotated in the reference.}
In all cases, more than half of the annotated words can be evaluated for gender translation.

Regarding gender translation accuracy, \texttt{enc-dec} provides the highest scores (avg. $85.57$), \df{while \texttt{ZeroSwot} and \texttt{Seamless} register lower average scores of $62.80$ and $53.35$, respectively.} 
Notably, speech+MT models with stronger overall translation quality exhibit lower accuracy. 
Instead, as shown in Figure \ref{fig:correlation}, accuracy scores strongly correlate with 
probing performance 
($R^2 = 0.99$, \textit{p} 
$< 0.01$). 
In other words,
\textbf{higher gender encoding capability leads to more accurate gender translation}, supporting the idea that ST models rely on acoustic cues to translate speaker-referred words when other signals are absent. 
When gender information is minimal, translations skew masculine. For instance, \texttt{Seamless}, for which gender encoding is almost absent  (low F1 scores in Table \ref{tab:results_probing}), strongly favors masculine forms, with feminine score only peaking at $17.60$  (en$\rightarrow$fr).
\df{A similar bias is observed in \texttt{ZeroSwot}, where average accuracy is $50.69$ for the \texttt{She} class and $74.90$ for the \texttt{He} class.
Even \texttt{enc-dec} exhibits a masculine skew: although it achieves higher feminine accuracy compared to the other models (avg. 78.53), this remains significantly lower than its masculine score (avg. 87.62), despite clear evidence of gender encoding (see Table \ref{tab:results_probing}).}
This suggests that an underlying linguistic bias sometimes overrides acoustic cues (see Appendix \ref{app:fine-grained} for some examples).

All in all, ST models show variable performance on gender translation, with gender encoding serving as a proxy for the agreement between translation and the speaker's gender.
When gender encoding is low, a masculine bias
emerges.

\section{Conclusion}

We investigated how diverse ST architectures encode speaker gender information. Using attention-based probes, we found that while traditional encoder-decoder models trained solely on ST data 
retain this information, the adapters in newer speech+MT architectures tend to erase it. 
Moreover, gender encoding is correlated with the ST model's ability to assign the correct grammatical gender to words referring to the speaker. This finding suggests that ST models can leverage acoustic gender information when available. When such information is weakly encoded, the models default to masculine translations more frequently.

\df{Our study sheds new light on how ST models encode and use gender information during translation, opening avenues for further research. For example, future work could explore which properties of adapters---or aspects of their training---contribute to the loss of gender information, and whether similar patterns 
emerge for acoustic and paralinguistic features
that may be weakened during the mapping to the text embedding space. These findings also inform strategies to mitigate biases such as undue masculine translations. One approach could involve modifying the configuration of the adapters to better preserve the notion of the speaker's gender---when appropriate---or integrating external knowledge 
to avoid misgendering.}

\section{Acknowledgments}

\df{The work of Marco Gaido was funded by the PNRR project FAIR -  Future AI Research (PE00000013), under the NRRP MUR program funded by the NextGenerationEU. The work of Luisa Bentivogli and Matteo Negri was supported by the European Union's Horizon research and innovation programme under grant agreement No 101135798, project Meetween (My Personal AI Mediator for Virtual MEETtings BetWEEN People).}
The work of André F.T. Martins and Giuseppe Attanasio was supported by EU's Horizon Europe Research and Innovation Actions (UTTER, contract 101070631), by the project DECOLLAGE (ERC-2022-CoG 101088763), by the Portuguese Recovery and Resilience Plan through project C645008882-00000055 (Center for Responsible AI), and by FCT/MECI through national funds and when applicable co-funded EU funds under UID/50008: Instituto de Telecomunicações. 

\section{Limitations}
\label{sec:limitations}

\paragraph{Models and Languages}
Our analysis focuses on specific models and language directions. The choice of models was driven by their performance and widespread use in the selected language pairs. The selection of languages was based on the need to translate gender-neutral forms into gender-marked ones and the availability of annotated ST data. 
As more data becomes available, we plan to expand our analysis to include additional translation systems and languages, as well as systems which integrate speech into Large Language Models \cite{gaido2024speechtranslationspeechfoundation}.
Nevertheless, prior research \cite{guillaume24_interspeech, attanasio-etal-2024-twists} suggests that gender encoding mechanisms are consistent across languages, so it is likely that similar patterns would emerge in other languages.

\paragraph{Probing} 
While the probing paradigm \cite{conneau-etal-2018-cram} is intuitive and well-established, it has been criticized for not directly confirming whether the model uses the extracted information \cite{probing_belinkov}. For example, some studies have highlighted a mismatch between probe performance and the original model’s performance in NLP tasks \cite{belinkov-glass-2019-analysis, ravichander-etal-2021-probing, elazar_etal_2021}.
Our experiments demonstrate a strong correlation between gender classification through probes (as an auxiliary task) and a specific translation aspect: gender assignment to words referring to the speaker (the original \textit{sub}task).
Although correlation does not imply causation, this finding proves that ST models likely use acoustic information for gender translation. To build on this, in future work, we plan to explore further interpretability techniques, such as \textit{amnesic probing} \cite{elazar_etal_2021} or 
\df{other methods in the field of mechanistic interpretability \cite{ferrando-etal-2024, saphra-wiegreffe-2024-mechanistic}.}

\section{Ethics Statement}
\label{sec:ethics_statement}

Following prior work \cite{deseyssel22_interspeech, guillaume24_interspeech, chowdhury-2024, krishnan24_interspeech, attanasio-etal-2024-twists},  we use the term ``gender'' as an approximation to account for documented differences in voices \textit{and} linguistic expression of gender identity (\texttt{She} or \texttt{He}).
Adopting this framework, we investigated
how ST models encode gender-related vocal differences, which may stem from both physiological factors---such as anatomical differences in vocal tracts between male and female speakers \cite{simpson_2001, hillenbrand-2009-pitch}---and sociocultural aspects---such as vocal behaviors associated with masculinity or femininity \cite{coleman-1976-voice, nylen_2004}.
Accordingly, to analyze how gender encoding capabilities affect ST systems' bias, we adopted a binary framework with only \texttt{She}/\texttt{He} class labels, primarily due to the lack of extensive speech data featuring non-binary voices. However, we recognize that vocal differences related to gender exist on a continuum, just as gender identities do. Expanding research to encompass non-binary identities is an essential next step.

\df{In the context of translation, \textit{gender bias} refers to the tendency of systems to favor one gender form over another---typically masculine over feminine---or to associate translations of specific roles and professions with a particular gender based on stereotypes \cite{savoldi-etal-2021-gender}. These biases can affect users' self-perception, as gendered language plays a fundamental role in shaping personal identity and representation \cite{stahlberg-gender, corbett, gygax}. As noted by \citet{blodgett-etal-2020-language} and \citet{savoldi-etal-2021-gender}, gender bias in translation technologies can lead to representational harms, such as reducing the visibility of women or reinforcing negative stereotypes about 
gender groups,
as well as allocational harms, including disparities in the quality of service received by male and female users.
Our experiments show that some models exhibit a strong bias toward masculine forms, posing a risk of such harms. We also find that encoding vocal characteristics related to gender can affect the accuracy of speaker gender translation.}

\df{Leveraging acoustic information was a key motivation behind the shift from cascaded systems---where vocal information was lost between ASR and MT---to direct models aimed at improving ST quality \cite{sperber-paulik-2020-speech, bentivogli-etal-2021-cascade}.}
\df{Our study provides a concrete example, demonstrating that accuracy of gender translation improves when gender-related vocal differences are encoded by the ST model. However, we do not necessarily advocate leveraging vocal features to assign linguistic gender markers, as such decisions may not align with a speaker’s gender identity. This concern is particularly relevant for transgender individuals, children, and people with vocal impairments \cite{matar_2016, pereira, VillasBoas, menezes2022}. To avoid the risk of misgendering in our evaluation, we relied on self-declared gender identities. Although vocal properties in real-world scenarios may lead to misgendering and should be interpreted with caution, we argue that our findings nonetheless provide valuable insights.}
\df{On one hand, our work contributes to advancing the understanding of gender bias in ST. On the other hand, raising awareness of how ST models encode speaker-related information---and how this influences translations---can help stakeholders make informed decisions to prevent harmful misgendering. It also empowers developers to make deliberate choices regarding model architecture and training strategies, as we found that specific architectures affect how gender information is encoded and used. Therefore, by enhancing the interpretability of ST models, we believe our work contributes to positive social impact.}

\bibliography{main}

\begin{thebibliography}{59}
\providecommand{\natexlab}[1]{#1}

\bibitem[{Attanasio et~al.(2024)Attanasio, Savoldi, Fucci, and Hovy}]{attanasio-etal-2024-twists}
Giuseppe Attanasio, Beatrice Savoldi, Dennis Fucci, and Dirk Hovy. 2024.
\newblock \href {https://doi.org/10.18653/v1/2024.emnlp-main.1188} {Twists, humps, and pebbles: Multilingual speech recognition models exhibit gender performance gaps}.
\newblock In \emph{Proceedings of the 2024 Conference on Empirical Methods in Natural Language Processing}, pages 21318--21340, Miami, Florida, USA. Association for Computational Linguistics.

\bibitem[{Azul(2013)}]{azul2013voices}
David Azul. 2013.
\newblock How do voices become gendered? a critical examination of everyday and medical constructions of the relationship between voice, sex, and gender identity.
\newblock In \emph{Challenging popular myths of sex, gender and biology}, pages 77--88. Springer.

\bibitem[{Baevski et~al.(2020)Baevski, Zhou, Mohamed, and Auli}]{wav2vec2.0}
Alexei Baevski, Henry Zhou, Abdelrahman Mohamed, and Michael Auli. 2020.
\newblock \href {https://arxiv.org/abs/2006.11477} {wav2vec 2.0: A framework for self-supervised learning of speech representations}.
\newblock \emph{Preprint}, arXiv:2006.11477.

\bibitem[{Belinkov(2022)}]{probing_belinkov}
Yonatan Belinkov. 2022.
\newblock \href {https://doi.org/10.1162/coli_a_00422} {Probing classifiers: Promises, shortcomings, and advances}.
\newblock \emph{Computational Linguistics}, 48(1):207--219.

\bibitem[{Belinkov and Glass(2019)}]{belinkov-glass-2019-analysis}
Yonatan Belinkov and James Glass. 2019.
\newblock \href {https://doi.org/10.1162/tacl_a_00254} {Analysis methods in neural language processing: A survey}.
\newblock \emph{Transactions of the Association for Computational Linguistics}, 7:49--72.

\bibitem[{Bentivogli et~al.(2021)Bentivogli, Cettolo, Gaido, Karakanta, Martinelli, Negri, and Turchi}]{bentivogli-etal-2021-cascade}
Luisa Bentivogli, Mauro Cettolo, Marco Gaido, Alina Karakanta, Alberto Martinelli, Matteo Negri, and Marco Turchi. 2021.
\newblock \href {https://doi.org/10.18653/v1/2021.acl-long.224} {{Cascade versus Direct Speech Translation: Do the Differences Still Make a Difference?}}
\newblock In \emph{Proceedings of the 59th Annual Meeting of the Association for Computational Linguistics and the 11th International Joint Conference on Natural Language Processing (Volume 1: Long Papers)}, pages 2873--2887, Online. Association for Computational Linguistics.

\bibitem[{Bentivogli et~al.(2020)Bentivogli, Savoldi, Negri, Di~Gangi, Cattoni, and Turchi}]{bentivogli-etal-2020-gender}
Luisa Bentivogli, Beatrice Savoldi, Matteo Negri, Mattia~A. Di~Gangi, Roldano Cattoni, and Marco Turchi. 2020.
\newblock \href {https://doi.org/10.18653/v1/2020.acl-main.619} {Gender in danger? evaluating speech translation technology on the {M}u{ST}-{SHE} corpus}.
\newblock In \emph{Proceedings of the 58th Annual Meeting of the Association for Computational Linguistics}, pages 6923--6933, Online. Association for Computational Linguistics.

\bibitem[{Blodgett et~al.(2020)Blodgett, Barocas, Daum{\'e}~III, and Wallach}]{blodgett-etal-2020-language}
Su~Lin Blodgett, Solon Barocas, Hal Daum{\'e}~III, and Hanna Wallach. 2020.
\newblock \href {https://doi.org/10.18653/v1/2020.acl-main.485} {Language (technology) is power: A critical survey of {``}bias{''} in {NLP}}.
\newblock In \emph{Proceedings of the 58th Annual Meeting of the Association for Computational Linguistics}, pages 5454--5476, Online. Association for Computational Linguistics.

\bibitem[{Cattoni et~al.(2021)Cattoni, {Di Gangi}, Bentivogli, Negri, and Turchi}]{mustc}
Roldano Cattoni, Mattia~Antonino {Di Gangi}, Luisa Bentivogli, Matteo Negri, and Marco Turchi. 2021.
\newblock \href {https://doi.org/10.1016/j.csl.2020.101155} {{MuST-C: A multilingual corpus for end-to-end speech translation}}.
\newblock \emph{Computer Speech \& Language}, 66:101155.

\bibitem[{Choi et~al.(2024)Choi, Pasad, Nakamura, Fukayama, Livescu, and Watanabe}]{choi24b_interspeech}
Kwanghee Choi, Ankita Pasad, Tomohiko Nakamura, Satoru Fukayama, Karen Livescu, and Shinji Watanabe. 2024.
\newblock \href {https://doi.org/10.21437/Interspeech.2024-1157} {Self-supervised speech representations are more phonetic than semantic}.
\newblock In \emph{Interspeech 2024}, pages 4578--4582.

\bibitem[{Chowdhury et~al.(2024)Chowdhury, Durrani, and Ali}]{chowdhury-2024}
Shammur~Absar Chowdhury, Nadir Durrani, and Ahmed Ali. 2024.
\newblock \href {https://doi.org/10.1016/j.csl.2023.101539} {What do end-to-end speech models learn about speaker, language and channel information? a layer-wise and neuron-level analysis}.
\newblock \emph{Computer Speech \& Language}, 83:101539.

\bibitem[{Chung et~al.(2021)Chung, Zhang, Han, Chiu, Qin, Pang, and Wu}]{w2vbert}
Yu-An Chung, Yu~Zhang, Wei Han, Chung-Cheng Chiu, James Qin, Ruoming Pang, and Yonghui Wu. 2021.
\newblock \href {https://arxiv.org/abs/2108.06209} {W2v-bert: Combining contrastive learning and masked language modeling for self-supervised speech pre-training}.
\newblock \emph{Preprint}, arXiv:2108.06209.

\bibitem[{Coleman(1976)}]{coleman-1976-voice}
Ralph~O. Coleman. 1976.
\newblock \href {https://doi.org/10.1044/jshr.1901.168} {A comparison of the contributions of two voice quality characteristics to the perception of maleness and femaleness in the voice}.
\newblock \emph{Journal of Speech \& Hearing Research}, 19(1):168--180.

\bibitem[{Communication et~al.(2023)Communication, Barrault, Chung, Meglioli, Dale et~al.}]{seamless}
Seamless Communication, Loïc Barrault, Yu-An Chung, Mariano~Cora Meglioli, David Dale, et~al. 2023.
\newblock \href {https://arxiv.org/abs/2308.11596} {{SeamlessM4T: Massively Multilingual \& Multimodal Machine Translation}}.
\newblock \emph{Preprint}, arXiv:2308.11596.

\bibitem[{Conneau et~al.(2018)Conneau, Kruszewski, Lample, Barrault, and Baroni}]{conneau-etal-2018-cram}
Alexis Conneau, German Kruszewski, Guillaume Lample, Lo{\"i}c Barrault, and Marco Baroni. 2018.
\newblock \href {https://doi.org/10.18653/v1/P18-1198} {What you can cram into a single {\$}{\&}!{\#}* vector: Probing sentence embeddings for linguistic properties}.
\newblock In \emph{Proceedings of the 56th Annual Meeting of the Association for Computational Linguistics (Volume 1: Long Papers)}, pages 2126--2136, Melbourne, Australia. Association for Computational Linguistics.

\bibitem[{Corbett(2013)}]{corbett}
Greville~G. Corbett. 2013.
\newblock \emph{The Expression of Gender}.
\newblock De Gruyter.

\bibitem[{Cormac~English et~al.(2022)Cormac~English, Kelleher, and Carson-Berndsen}]{cormac-english-etal-2022-domain}
Patrick Cormac~English, John~D. Kelleher, and Julie Carson-Berndsen. 2022.
\newblock \href {https://doi.org/10.18653/v1/2022.sigmorphon-1.9} {Domain-informed probing of wav2vec 2.0 embeddings for phonetic features}.
\newblock In \emph{Proceedings of the 19th SIGMORPHON Workshop on Computational Research in Phonetics, Phonology, and Morphology}, pages 83--91, Seattle, Washington. Association for Computational Linguistics.

\bibitem[{{de Seyssel} et~al.(2022){de Seyssel}, Lavechin, Adi, Dupoux, and Wisniewski}]{deseyssel22_interspeech}
Maureen {de Seyssel}, Marvin Lavechin, Yossi Adi, Emmanuel Dupoux, and Guillaume Wisniewski. 2022.
\newblock \href {https://doi.org/10.21437/Interspeech.2022-373} {Probing phoneme, language and speaker information in unsupervised speech representations}.
\newblock In \emph{Interspeech 2022}, pages 1402--1406.

\bibitem[{Elazar et~al.(2021)Elazar, Ravfogel, Jacovi, and Goldberg}]{elazar_etal_2021}
Yanai Elazar, Shauli Ravfogel, Alon Jacovi, and Yoav Goldberg. 2021.
\newblock \href {https://doi.org/10.1162/tacl_a_00359} {Amnesic probing: Behavioral explanation with amnesic counterfactuals}.
\newblock \emph{Transactions of the Association for Computational Linguistics}, 9:160--175.

\bibitem[{Ferrando et~al.(2024)Ferrando, Sarti, Bisazza, and Costa-jussà}]{ferrando-etal-2024}
Javier Ferrando, Gabriele Sarti, Arianna Bisazza, and Marta~R. Costa-jussà. 2024.
\newblock \href {https://arxiv.org/abs/2405.00208} {{A Primer on the Inner Workings of Transformer-based Language Models}}.
\newblock \emph{Preprint}, arXiv:2405.00208.

\bibitem[{Frogner et~al.(2015)Frogner, Zhang, Mobahi, Araya-Polo, and Poggio}]{frogner-2015}
Charlie Frogner, Chiyuan Zhang, Hossein Mobahi, Mauricio Araya-Polo, and Tomaso Poggio. 2015.
\newblock Learning with a wasserstein loss.
\newblock In \emph{Proceedings of the 29th International Conference on Neural Information Processing Systems - Volume 2}, NIPS'15, page 2053–2061, Cambridge, MA, USA. MIT Press.

\bibitem[{Fuchs and Toda(2010)}]{fuchs2010differences}
Susanne Fuchs and Martine Toda. 2010.
\newblock Do differences in male versus female/s/reflect biological or sociophonetic factors.
\newblock \emph{Turbulent sounds: An interdisciplinary guide}, 21:281--302.

\bibitem[{Gaido et~al.(2024)Gaido, Papi, Negri, and Bentivogli}]{gaido2024speechtranslationspeechfoundation}
Marco Gaido, Sara Papi, Matteo Negri, and Luisa Bentivogli. 2024.
\newblock \href {https://arxiv.org/abs/2402.12025} {{Speech Translation with Speech Foundation Models and Large Language Models: What is There and What is Missing?}}
\newblock \emph{Preprint}, arXiv:2402.12025.

\bibitem[{Gaido et~al.(2020)Gaido, Savoldi, Bentivogli, Negri, and Turchi}]{gaido-etal-2020-breeding}
Marco Gaido, Beatrice Savoldi, Luisa Bentivogli, Matteo Negri, and Marco Turchi. 2020.
\newblock \href {https://doi.org/10.18653/v1/2020.coling-main.350} {Breeding gender-aware direct speech translation systems}.
\newblock In \emph{Proceedings of the 28th International Conference on Computational Linguistics}, pages 3951--3964, Barcelona, Spain (Online). International Committee on Computational Linguistics.

\bibitem[{Glorot and Bengio(2010)}]{xavier}
Xavier Glorot and Yoshua Bengio. 2010.
\newblock \href {https://proceedings.mlr.press/v9/glorot10a.html} {Understanding the difficulty of training deep feedforward neural networks}.
\newblock In \emph{Proceedings of the Thirteenth International Conference on Artificial Intelligence and Statistics}, volume~9 of \emph{Proceedings of Machine Learning Research}, pages 249--256, Chia Laguna Resort, Sardinia, Italy. PMLR.

\bibitem[{Guillaume et~al.(2024)Guillaume, Fily, Michaud, and Wisniewski}]{guillaume24_interspeech}
Séverine Guillaume, Maxime Fily, Alexis Michaud, and Guillaume Wisniewski. 2024.
\newblock \href {https://doi.org/10.21437/Interspeech.2024-953} {Gender and language identification in multilingual models of speech: Exploring the genericity and robustness of speech representations}.
\newblock In \emph{Interspeech 2024}, pages 3330--3334.

\bibitem[{Gygax et~al.(2019)Gygax, Elmiger, Zufferey, Garnham, Sczesny, von Stockhausen, and Oakhill}]{gygax}
Pascal~M. Gygax, Daniel Elmiger, Sandrine Zufferey, Alan Garnham, Sabine Sczesny, Friederike von Stockhausen, Lisa~Braun, and Jane Oakhill. 2019.
\newblock \href {https://doi.org/10.3389/fpsyg.2019.01604} {A language index of grammatical gender dimensions to study the impact of grammatical gender on the way we perceive women and men}.
\newblock \emph{Frontiers in Psychology}, 10.

\bibitem[{Hewitt and Liang(2019)}]{hewitt-liang-2019-designing}
John Hewitt and Percy Liang. 2019.
\newblock \href {https://doi.org/10.18653/v1/D19-1275} {Designing and interpreting probes with control tasks}.
\newblock In \emph{Proceedings of the 2019 Conference on Empirical Methods in Natural Language Processing and the 9th International Joint Conference on Natural Language Processing (EMNLP-IJCNLP)}, pages 2733--2743, Hong Kong, China. Association for Computational Linguistics.

\bibitem[{Hillenbrand and Clark(2009)}]{hillenbrand-2009-pitch}
James~M. Hillenbrand and Michael~J. Clark. 2009.
\newblock \href {https://doi.org/10.3758/APP.71.5.1150} {The role of f0 and formant frequencies in distinguishing the voices of men and women}.
\newblock \emph{Attention Perception \& Psychophysics}, 71(5):1150--1166.

\bibitem[{Krishnan et~al.(2024)Krishnan, Abdullah, and Klakow}]{krishnan24_interspeech}
Aravind Krishnan, Badr~M. Abdullah, and Dietrich Klakow. 2024.
\newblock \href {https://doi.org/10.21437/Interspeech.2024-2209} {On the encoding of gender in transformer-based asr representations}.
\newblock In \emph{Interspeech 2024}, pages 3090--3094.

\bibitem[{Leung et~al.(2018)Leung, Oates, and Chan}]{leung2018voice}
Yeptain Leung, Jennifer Oates, and Siew~Pang Chan. 2018.
\newblock Voice, articulation, and prosody contribute to listener perceptions of speaker gender: A systematic review and meta-analysis.
\newblock \emph{Journal of Speech, Language, and Hearing Research}, 61(2):266--297.

\bibitem[{Li et~al.(2023)Li, Li, Savarese, and Hoi}]{qformer}
Junnan Li, Dongxu Li, Silvio Savarese, and Steven Hoi. 2023.
\newblock Blip-2: bootstrapping language-image pre-training with frozen image encoders and large language models.
\newblock In \emph{Proceedings of the 40th International Conference on Machine Learning}, ICML'23. JMLR.org.

\bibitem[{Matar et~al.(2016)Matar, Portes, Lancia, Legou, and Baider}]{matar_2016}
Nayla Matar, Cristel Portes, Leonardo Lancia, Thierry Legou, and Fabienne Baider. 2016.
\newblock \href {https://hal.archives-ouvertes.fr/hal-01459619} {{Voice quality and gender stereotypes: A study on Lebanese women with Reinke's edema}}.
\newblock \emph{{Journal of Speech, Language, and Hearing Research}}, 59(6):1608--1617.

\bibitem[{Menezes et~al.(2022)Menezes, {de Lira}, de~Araújo, {de Almeida}, de~Oliveira Camargo~Gomes, Moraes, and Lucena}]{menezes2022}
Danielle~Pereira Menezes, Zulina~Souza {de Lira}, Ana Nery~Barbosa de~Araújo, Anna Alice~Figueirêdo {de Almeida}, Adriana de~Oliveira Camargo~Gomes, Bruno~Teixeira Moraes, and Jonia~Alves Lucena. 2022.
\newblock \href {https://doi.org/10.1016/j.jvoice.2021.12.020} {Prosodic differences in the voices of transgender and cisgender women: Self-perception of voice - an auditory and acoustic analysis}.
\newblock \emph{Journal of Voice}.

\bibitem[{Nylén et~al.(2024)Nylén, Holmberg, and Södersten}]{nylen_2004}
Frida Nylén, Johan Holmberg, and Maria Södersten. 2024.
\newblock \href {https://doi.org/10.1121/10.0025932} {Acoustic cues to femininity and masculinity in spontaneous speech}.
\newblock \emph{The Journal of the Acoustical Society of America}, 155(5):3090--3100.

\bibitem[{Pasad et~al.(2023)Pasad, Shi, and Livescu}]{pasad_etal_2023}
Ankita Pasad, Bowen Shi, and Karen Livescu. 2023.
\newblock \href {https://doi.org/10.1109/ICASSP49357.2023.10096149} {Comparative layer-wise analysis of self-supervised speech models}.
\newblock In \emph{ICASSP 2023 - 2023 IEEE International Conference on Acoustics, Speech and Signal Processing (ICASSP)}, pages 1--5.

\bibitem[{Paszke et~al.(2019)Paszke, Gross, Massa, Lerer, Bradbury, Chanan, Killeen, Lin, Gimelshein, Antiga, Desmaison, K\"{o}pf, Yang, DeVito, Raison, Tejani, Chilamkurthy, Steiner, Fang, Bai, and Chintala}]{pytorch}
Adam Paszke, Sam Gross, Francisco Massa, Adam Lerer, James Bradbury, Gregory Chanan, Trevor Killeen, Zeming Lin, Natalia Gimelshein, Luca Antiga, Alban Desmaison, Andreas K\"{o}pf, Edward Yang, Zach DeVito, Martin Raison, Alykhan Tejani, Sasank Chilamkurthy, Benoit Steiner, Lu~Fang, Junjie Bai, and Soumith Chintala. 2019.
\newblock Pytorch: an imperative style, high-performance deep learning library.
\newblock In \emph{Proceedings of the 33rd International Conference on Neural Information Processing Systems}. Curran Associates Inc., Red Hook, NY, USA.

\bibitem[{Pedregosa et~al.(2011)Pedregosa, Varoquaux, Gramfort, Michel, Thirion, Grisel, Blondel, Prettenhofer, Weiss, Dubourg, Vanderplas, Passos, Cournapeau, Brucher, Perrot, and Duchesnay}]{scikitlearn}
Fabian Pedregosa, Ga\"{e}l Varoquaux, Alexandre Gramfort, Vincent Michel, Bertrand Thirion, Olivier Grisel, Mathieu Blondel, Peter Prettenhofer, Ron Weiss, Vincent Dubourg, Jake Vanderplas, Alexandre Passos, David Cournapeau, Matthieu Brucher, Matthieu Perrot, and \'{E}douard Duchesnay. 2011.
\newblock Scikit-learn: Machine learning in python.
\newblock \emph{J. Mach. Learn. Res.}, 12(null):2825–2830.

\bibitem[{Pereira et~al.(2018)Pereira, Dassie-Leite, Pereira, Cavichiolo, Rosa, and Fugmann}]{pereira}
Amanda~Maria Pereira, Ana~Paula Dassie-Leite, Eliane~Cristina Pereira, Juliana~Benthien Cavichiolo, Marcelo de~Oliveira Rosa, and Elmar~Allen Fugmann. 2018.
\newblock \href {https://doi.org/10.1590/2317-1782/20182017046} {Auditory perception of lay judges about gender identification of women with reinke's edema}.
\newblock \emph{CoDAS}, 30(4).

\bibitem[{Peyré and Cuturi(2019)}]{optimal-transport}
Gabriel Peyré and Marco Cuturi. 2019.
\newblock \href {https://doi.org/10.1561/2200000073} {Computational optimal transport: With applications to data science}.
\newblock \emph{Foundations and Trends® in Machine Learning}, 11(5-6):355--607.

\bibitem[{Prasad and Jyothi(2020)}]{prasad-jyothi-2020-accents}
Archiki Prasad and Preethi Jyothi. 2020.
\newblock \href {https://doi.org/10.18653/v1/2020.acl-main.345} {How accents confound: Probing for accent information in end-to-end speech recognition systems}.
\newblock In \emph{Proceedings of the 58th Annual Meeting of the Association for Computational Linguistics}, pages 3739--3753, Online. Association for Computational Linguistics.

\bibitem[{Ravichander et~al.(2021)Ravichander, Belinkov, and Hovy}]{ravichander-etal-2021-probing}
Abhilasha Ravichander, Yonatan Belinkov, and Eduard Hovy. 2021.
\newblock \href {https://doi.org/10.18653/v1/2021.eacl-main.295} {Probing the probing paradigm: Does probing accuracy entail task relevance?}
\newblock In \emph{Proceedings of the 16th Conference of the European Chapter of the Association for Computational Linguistics: Main Volume}, pages 3363--3377, Online. Association for Computational Linguistics.

\bibitem[{Rei et~al.(2020)Rei, Stewart, Farinha, and Lavie}]{rei-etal-2020-comet}
Ricardo Rei, Craig Stewart, Ana~C Farinha, and Alon Lavie. 2020.
\newblock \href {https://doi.org/10.18653/v1/2020.emnlp-main.213} {{COMET}: A neural framework for {MT} evaluation}.
\newblock In \emph{Proceedings of the 2020 Conference on Empirical Methods in Natural Language Processing (EMNLP)}, pages 2685--2702, Online. Association for Computational Linguistics.

\bibitem[{Saphra and Wiegreffe(2024)}]{saphra-wiegreffe-2024-mechanistic}
Naomi Saphra and Sarah Wiegreffe. 2024.
\newblock \href {https://doi.org/10.18653/v1/2024.blackboxnlp-1.30} {Mechanistic?}
\newblock In \emph{Proceedings of the 7th BlackboxNLP Workshop: Analyzing and Interpreting Neural Networks for NLP}, pages 480--498, Miami, Florida, US. Association for Computational Linguistics.

\bibitem[{Savoldi et~al.(2021)Savoldi, Gaido, Bentivogli, Negri, and Turchi}]{savoldi-etal-2021-gender}
Beatrice Savoldi, Marco Gaido, Luisa Bentivogli, Matteo Negri, and Marco Turchi. 2021.
\newblock \href {https://doi.org/10.1162/tacl_a_00401} {Gender bias in machine translation}.
\newblock \emph{Transactions of the Association for Computational Linguistics}, 9:845--874.

\bibitem[{Savoldi et~al.(2022)Savoldi, Gaido, Bentivogli, Negri, and Turchi}]{savoldi-etal-2022-morphosyntactic}
Beatrice Savoldi, Marco Gaido, Luisa Bentivogli, Matteo Negri, and Marco Turchi. 2022.
\newblock \href {https://doi.org/10.18653/v1/2022.acl-long.127} {Under the morphosyntactic lens: A multifaceted evaluation of gender bias in speech translation}.
\newblock In \emph{Proceedings of the 60th Annual Meeting of the Association for Computational Linguistics (Volume 1: Long Papers)}, pages 1807--1824, Dublin, Ireland. Association for Computational Linguistics.

\bibitem[{Shen et~al.(2024)Shen, Watkins, Alishahi, Bisazza, and Chrupa{\l}a}]{shen-etal-2024-encoding}
Gaofei Shen, Michaela Watkins, Afra Alishahi, Arianna Bisazza, and Grzegorz Chrupa{\l}a. 2024.
\newblock \href {https://doi.org/10.18653/v1/2024.naacl-long.239} {Encoding of lexical tone in self-supervised models of spoken language}.
\newblock In \emph{Proceedings of the 2024 Conference of the North American Chapter of the Association for Computational Linguistics: Human Language Technologies (Volume 1: Long Papers)}, pages 4250--4261, Mexico City, Mexico. Association for Computational Linguistics.

\bibitem[{Simpson(2001)}]{simpson_2001}
Adrian~P. Simpson. 2001.
\newblock \href {https://doi.org/10.1121/1.1356020} {Dynamic consequences of differences in male and female vocal tract dimensions}.
\newblock \emph{The Journal of the Acoustical Society of America}, 109(5 Pt 1):2153--2164.

\bibitem[{Sperber and Paulik(2020)}]{sperber-paulik-2020-speech}
Matthias Sperber and Matthias Paulik. 2020.
\newblock \href {https://doi.org/10.18653/v1/2020.acl-main.661} {Speech translation and the end-to-end promise: Taking stock of where we are}.
\newblock In \emph{Proceedings of the 58th Annual Meeting of the Association for Computational Linguistics}, pages 7409--7421, Online. Association for Computational Linguistics.

\bibitem[{Stahlberg et~al.(2007)Stahlberg, Braun, Irmen, and Sczesny}]{stahlberg-gender}
Dagmar Stahlberg, Friederike Braun, Lisa Irmen, and Sabine Sczesny. 2007.
\newblock \href {https://psycnet.apa.org/record/2007-01308-006} {Representation of the sexes in language}.
\newblock In Klaus Fiedler, editor, \emph{Social Communication}, pages 163--187. Psychology Press.

\bibitem[{Sun et~al.(2019)Sun, Gaut, Tang, Huang, ElSherief, Zhao, Mirza, Belding, Chang, and Wang}]{sun-etal-2019-mitigating}
Tony Sun, Andrew Gaut, Shirlyn Tang, Yuxin Huang, Mai ElSherief, Jieyu Zhao, Diba Mirza, Elizabeth Belding, Kai-Wei Chang, and William~Yang Wang. 2019.
\newblock \href {https://doi.org/10.18653/v1/P19-1159} {Mitigating gender bias in natural language processing: Literature review}.
\newblock In \emph{Proceedings of the 57th Annual Meeting of the Association for Computational Linguistics}, pages 1630--1640, Florence, Italy. Association for Computational Linguistics.

\bibitem[{Team et~al.(2022)Team, Costa-jussà, Cross, Çelebi et~al.}]{nllb}
NLLB Team, Marta~R. Costa-jussà, James Cross, Onur Çelebi, et~al. 2022.
\newblock \href {https://arxiv.org/abs/2207.04672} {No language left behind: Scaling human-centered machine translation}.
\newblock \emph{Preprint}, arXiv:2207.04672.

\bibitem[{Tsiamas et~al.(2024)Tsiamas, G{\'a}llego, Fonollosa, and Costa-juss{\`a}}]{tsiamas-etal-2024-pushing}
Ioannis Tsiamas, Gerard G{\'a}llego, Jos{\'e} Fonollosa, and Marta Costa-juss{\`a}. 2024.
\newblock \href {https://doi.org/10.18653/v1/2024.findings-acl.847} {Pushing the limits of zero-shot end-to-end speech translation}.
\newblock In \emph{Findings of the Association for Computational Linguistics: ACL 2024}, pages 14245--14267, Bangkok, Thailand. Association for Computational Linguistics.

\bibitem[{Villas-Bôas et~al.(2021)Villas-Bôas, Schwarz, Fontanari, Costa, Cardoso~da Silva, Schneider, Cielo, Spritzer, and Rodrigues~Lobato}]{VillasBoas}
Anna~Paula Villas-Bôas, Karine Schwarz, Anna Martha~Vaitses Fontanari, Angelo~Brandelli Costa, Dhiordan Cardoso~da Silva, Maiko~Abel Schneider, Carla~Aparecida Cielo, Poli~Mara Spritzer, and Maria~Inês Rodrigues~Lobato. 2021.
\newblock \href {https://doi.org/10.3389/fpsyg.2021.622526} {Acoustic measures of brazilian transgender women's voices: A case–control study}.
\newblock \emph{Frontiers in Psychology}, 12.

\bibitem[{Waheed et~al.(2024)Waheed, Atwany, Raj, and Singh}]{waheed2024speechfoundationmodelslearn}
Abdul Waheed, Hanin Atwany, Bhiksha Raj, and Rita Singh. 2024.
\newblock \href {https://arxiv.org/abs/2410.12948} {What do speech foundation models not learn about speech?}
\newblock \emph{Preprint}, arXiv:2410.12948.

\bibitem[{Wang et~al.(2020)Wang, Tang, Ma, Wu, Okhonko, and Pino}]{wang2020fairseqs2t}
Changhan Wang, Yun Tang, Xutai Ma, Anne Wu, Dmytro Okhonko, and Juan Pino. 2020.
\newblock fairseq s2t: Fast speech-to-text modeling with fairseq.
\newblock In \emph{Proceedings of the 2020 Conference of the Asian Chapter of the Association for Computational Linguistics (AACL): System Demonstrations}.

\bibitem[{Wolf et~al.(2020)Wolf, Debut, Sanh, Chaumond, Delangue, Moi, Cistac, Rault, Louf, Funtowicz, Davison, Shleifer, von Platen, Ma, Jernite, Plu, Xu, Le~Scao, Gugger, Drame, Lhoest, and Rush}]{wolf-etal-2020-transformers}
Thomas Wolf, Lysandre Debut, Victor Sanh, Julien Chaumond, Clement Delangue, Anthony Moi, Pierric Cistac, Tim Rault, Remi Louf, Morgan Funtowicz, Joe Davison, Sam Shleifer, Patrick von Platen, Clara Ma, Yacine Jernite, Julien Plu, Canwen Xu, Teven Le~Scao, Sylvain Gugger, Mariama Drame, Quentin Lhoest, and Alexander Rush. 2020.
\newblock \href {https://doi.org/10.18653/v1/2020.emnlp-demos.6} {Transformers: State-of-the-art natural language processing}.
\newblock In \emph{Proceedings of the 2020 Conference on Empirical Methods in Natural Language Processing: System Demonstrations}, pages 38--45, Online. Association for Computational Linguistics.

\bibitem[{Yang et~al.(2023)Yang, Shekar, Kang, and Hansen}]{yang23v_interspeech}
Mu~Yang, Ram C. M.~C. Shekar, Okim Kang, and John H.~L. Hansen. 2023.
\newblock \href {https://doi.org/10.21437/Interspeech.2023-2254} {What can an accent identifier learn? probing phonetic and prosodic information in a wav2vec2-based accent identification model}.
\newblock In \emph{Interspeech 2023}, pages 1923--1927.

\bibitem[{Zhao et~al.(2022)Zhao, Yang, Haffari, and Shareghi}]{zhao22g_interspeech}
Jinming Zhao, Hao Yang, Gholamreza Haffari, and Ehsan Shareghi. 2022.
\newblock \href {https://doi.org/10.21437/Interspeech.2022-592} {M-adapter: Modality adaptation for end-to-end speech-to-text translation}.
\newblock In \emph{Interspeech 2022}, pages 111--115.

\end{thebibliography}

\appendix

\begin{table*}[t!]
\centering
\footnotesize
\setlength{\tabcolsep}{5pt}
\begin{tabular}{l||c|c|c|c}
    \toprule
    & \textbf{\texttt{Seamless pre-ad}} &  \textbf{\texttt{ZeroSwot pre-ad}} & \textbf{\texttt{Enc-Dec}} & avg. \\
    \midrule
    \texttt{max pooling} & 61.37 & 54.46 & 65.48 & 60.44 \\
    \texttt{mean pooling} & 62.17 & 76.44 & 90.10 & 76.24 \\
    \texttt{positional sampling} & 71.45 & 64.96 & 83.37 & 73.26 \\
    \texttt{attention-based} & \textbf{75.53} & \textbf{89.60} & \textbf{92.21} & \textbf{85.78} \\
    \bottomrule
\end{tabular}
\caption{F1 scores of probes on \texttt{test-generic} across different ST model states and probing strategies.}
\label{tab:performance_comparison}
\end{table*}

\section{Probe Comparison}
\label{app:probe_comparison}

In our experiments, we use an attention-based probe, introduced in \S \ref{sec:gender_probing}. It is designed to predict a single label for each input sequence while preserving gender encoding across the temporal dimension. Unlike traditional approaches, our method avoids sampling or pooling mechanisms, which can undermine classification performance.

To assess its effectiveness, we compare our \texttt{attention-based} probe with three probing methods from prior work: logistic classifiers trained on hidden representations aggregated \textit{i)} via \texttt{max} pooling, \textit{ii)} via \texttt{mean} pooling, and \textit{iii)} \texttt{positional sampling} of hidden states at relative positions. In the latter case, we sample hidden states at every \df{$25\%$} of the sequence length and train five separate probes, each corresponding to one of these positions.
\df{We train these probes on the hidden states extracted from \texttt{Seamless/ZeroSwot pre-ad} and \texttt{enc-dec}, as these representations encode the speaker’s gender to a meaningful---although variable---extent. Table~\ref{tab:performance_comparison} reports the corresponding F1 scores on \texttt{test-generic}.}
For positional sampling, we report the highest scores among the five positions, which consistently correspond to the first position. Interestingly, this aligns with our analysis in Appendix \ref{app:gender_in_seq_len}, where we show that attention weight distributions tend to concentrate at early positions.

Our results show that the \texttt{attention-based} probe consistently outperforms all other probing strategies across all considered settings. Notably, \texttt{max pooling} performs the worst (avg. $60.44$), while \texttt{mean pooling} is the closest to the \texttt{attention-based} solution (avg. $76.24$ vs. $85.78$, respectively). The \texttt{positional sampling} method falls between \texttt{mean pooling} and \texttt{attention-based}, even 
though it surpasses
\texttt{mean pooling} on \texttt{Seamless pre-ad}.

Overall, our approach proves to be more effective than traditional methods in our scenario.
Moreover, our probe offers the additional advantage of interpretability by providing attention weights that reveal how gender information is distributed across the sequence
\df{(see Appendix \ref{app:gender_in_seq_len})}.

\section{Experimental Details}
\label{app:setting}

\paragraph{ST Models}
Below, we describe the architectures of the models used in our study (see \S \ref{sec:exp_setup}), along with the details for running inference operations and training the probes.
\begin{itemize}
\item Transformer-based encoder-decoder model \cite{wang2020fairseqs2t}: it employs a convolutional downsampler to reduce the length of speech inputs, followed by a Transformer encoder---initialized from an ASR model---and a decoder. The encoder output is used for the \texttt{enc-dec} setting. Both the encoder and decoder are jointly trained for the ST task using autoregressive cross-entropy loss.
\item SeamlessM4T \cite{seamless}: it uses a speech encoder based on the pre-trained w2v-BERT 2.0 \cite{w2vbert}, whose output is used in the \texttt{Seamless pre-ad} setting. A length adapter, derived from the M-adaptor \cite{zhao22g_interspeech}, processes the speech encoder's hidden states, and the resulting output is taken for the \texttt{Seamless post-ad} setting.
The speech encoder 
\df{followed by the length adapter} 
is paired with a text encoder based on the pre-trained No Language Left Behind (NLLB) encoder \cite{nllb}. Both encoders share a common text decoder, which is initialized from the NLLB decoder. The entire model is jointly trained using cross-entropy loss for speech-to-text and text-to-text translation tasks, along with token-level knowledge distillation from MT task (teacher) to the ST task (student).
\item ZeroSwot \cite{tsiamas-etal-2024-pushing}: it encodes speech using Wav2Vec 2.0 \cite{wav2vec2.0}, with its representations used in the \texttt{ZeroSwot pre-ad} setting. A CTC module then predicts characters from these representations. The CTC probabilities and Wav2Vec features are processed through an adapter to produce a compressed acoustic representation.
This representation is then 
enhanced with positional encodings to form the final speech embedding, which is used in the \texttt{ZeroSwot post-ad} setting. 
The Wav2Vec encoder, CTC module, and adapter are fine-tuned 
\df{by jointly minimizing the CTC loss and the Wasserstein loss \cite{frogner-2015} using Optimal Transport \cite{optimal-transport}, to align speech-derived representations with the text-derived embeddings produced by the NLLB encoder.}
\end{itemize}
Audio data are normalized to \texttt{float32} and truncated at $60$s.
Inference for all models is conducted using \texttt{transformers 4.47.0} \cite{wolf-etal-2020-transformers} on an NVIDIA A40 GPU ($48$GB RAM).
We use publicly available configuration files and model checkpoints from the HuggingFace Hub:
\url{https://huggingface.co/facebook/seamless-m4t-v2-large} for SeamlessM4T, \url{https://huggingface.co/johntsi/ZeroSwot-Large\_asr-mustc\_mt-mustc\_en-to-8} for ZeroSwot, \url{https://huggingface.co/facebook/s2t-medium-mustc-multilingual-st} for the Transformer-based encoder-decoder.

\paragraph{Probes} 
We implemented our attention-based probes using \texttt{PyTorch 2.3.0} \cite{pytorch}.
For training, 
we use a batch size of $32$ and a starting learning rate of $0.0001$, reduced by half if the validation loss does not decrease for $3$ consecutive epochs.
The weights of the classification layer are initialized using Xavier initialization \cite{xavier}. We use the Adam optimizer and optimize the model with cross-entropy loss.
Training stops if the validation loss on the \texttt{dev} set does not improve by at least $0.00001$ for $20$ epochs.
All training and testing were conducted on an NVIDIA Tesla K80 GPU ($12$GB RAM).
For the probes used in Appendix \ref{app:probe_comparison}, we used logistic classifier implemented with \texttt{scikit-learn 1.4.1} \cite{scikitlearn} (\texttt{SGDClassifier} class with \texttt{log loss} and default values).

\section{\df{Manual Evaluation}}
\label{sec:manual_eval}

\df{For the evaluation of gender translation, we adopt the official metrics from MuST-SHE \cite{bentivogli-etal-2020-gender}---\textit{gender accuracy} and \textit{coverage} (see \S\ref{sec:exp_setup}). 
\ga{Coverage is the percentage of sentences for which accuracy can be assessed. This number relies on automatic string matching, i.e., by checking whether the term annotated for gender appears, in any gendered form, in the system output. Accuracy is then measured only on the \textit{in-coverage} \df{(IC)} outputs by verifying the gender marking of the word found.}
Therefore, \ga{coverage is} complementary \ga{to accuracy}.}
\df{However, IC accuracy offers only a partial view, as \textit{out-of-coverage} (OOC) instances---excluded from gender accuracy calculations---can still convey gender information, for example, when the generated term is a gendered synonym. 
To assess whether OOC instances exhibit trends that align with or diverge from the accuracy scores observed in IC cases, one of the authors manually analyzed the OOC outputs for Italian translations produced by the three ST models.
}

\df{The manual analysis reveals that the OOC cases include: 
\textit{(i)} translations that avoid speaker-referred gendered words by using paraphrases---which can sometimes be incorrect---or epicene terms; \textit{(ii)} gendered synonyms; and \textit{(iii)} gendered mistranslations referring to the speaker. Gender accuracy evaluation is possible only for cases of type \textit{(ii)} and \textit{(iii)}. Examples of all three categories are provided below.}

\begin{quote}
\textbf{(i)} \\
\textbf{En. source}: \textit{And to make sure} [\ldots] \\
\textbf{It. reference}: \textit{Per essere \underline{sicuro}}\textsuperscript{M} \textit{che} [\ldots]  \\
\textbf{\texttt{Seamless} output}: \textit{E per \underline{assicurarmi} che} [\ldots] \\
\textbf{Note}: 
The model uses the verb ``assicurarmi'' (En. ``to make sure'') instead of the expression ``essere sicuro'' (En. ``to be sure'') found in the reference.

\vspace{0.8em}

\textbf{(ii)} \\
\textbf{En. source}: [\ldots] \textit{I was amazed} [\ldots] \\
\textbf{It. reference}: [\ldots] \textit{Ero \underline{stupefatto}}\textsuperscript{M}. [\ldots] \\
\textbf{\texttt{ZeroSwot} output}: [\ldots] \textit{Ero \underline{stupito}}\textsuperscript{M}. [\ldots] \\
\textbf{Note}: 
The adjective ``stupito'' is a synonym of ``stupefatto'', both meaning ``amazed''.

\vspace{0.8em}

\textbf{(iii)} \\
\textbf{En. source}: \textit{I'm exhausted, and I'm numb.} \\
\textbf{It. reference}: \textit{Sono \underline{esausta}}\textsuperscript{F}\textit{, e sono \underline{paralizzata}}\textsuperscript{F}. \\
\textbf{\texttt{Seamless} output}: \textit{Sono \underline{esausta}}\textsuperscript{F} \textit{e sono \underline{sordo}}\textsuperscript{M}. \\
\textbf{Note}: 
The adjective ``sordo'' (En. ``deaf'') is a mistranslation of ``numb'' in this context.
\end{quote}

\df{Cases of type \textit{(ii)} and \textit{(iii)} account for $48.40\%$ of the OOC instances for \texttt{Seamless}, $43.17\%$ for \texttt{ZeroSwot}, and $64.64\%$ for \texttt{enc-dec}. We incorporate these cases into the IC set and recompute the scores.
Table~\ref{tab:revised_accuracy} reports both the scores obtained using string-matching evaluation---as in Table~\ref{tab:performance}---and those combining string matching with manual assessment of OOC instances. Coverage increases in the latter due to the reduced number of unassessable cases. 
More importantly, accuracy scores remain largely consistent. Absolute differences between string-matching scores and those including manual evaluation range from just $0.23$ (for \texttt{ZeroSwot}, \texttt{She} class) to $2.89$ (for \texttt{ZeroSwot}, \texttt{He} class).
This consistency confirms that gender accuracy, when extended to include manually assessed OOC instances, aligns closely with that observed for IC cases, echoing findings by \citet{savoldi-etal-2022-morphosyntactic} in their analysis of other systems.}

\df{Overall, this manual evaluation suggests that, despite the limitations of string-matching metrics, the trends they reveal remain broadly reliable, and the minor variations in gender accuracy introduced by manual evaluation do not affect the main findings.}

\begin{table}[t]
    \footnotesize
    \centering
    \begin{tabular}{l|cc|cc}
    \toprule
    & \multicolumn{2}{c|}{\textbf{\texttt{She}}} & \multicolumn{2}{c}{\textbf{\texttt{He}}} \\
     & \textbf{Cov.} & \textbf{Acc.} & \textbf{Cov.} & \textbf{Acc.} \\
    \midrule
    & \multicolumn{4}{c}{\textit{String Matching}} \\
    \midrule
    \texttt{Seamless} & 57.11 & \cellcolor{gray!20}13.30 & 57.52 & \cellcolor{gray!20}91.36 \\
    \texttt{ZeroSwot} & 65.23 & \cellcolor{gray!20}46.18 & 60.19 & \cellcolor{gray!20}77.87 \\
    \texttt{enc-dec}    & 55.33 & \cellcolor{gray!20}76.44 & 54.85 & \cellcolor{gray!20}94.47 \\
    \midrule
        & \multicolumn{4}{c}{\textit{String Matching + Manual}} \\
    \midrule
    \texttt{Seamless} & 70.56 & \cellcolor{gray!20}12.59 & 73.79 & \cellcolor{gray!20}93.75 \\
    \texttt{ZeroSwot} & 78.43 & \cellcolor{gray!20}45.95 & 76.94 & \cellcolor{gray!20}80.76 \\
    \texttt{enc-dec}    & 71.57 & \cellcolor{gray!20}78.55 & 70.39 & \cellcolor{gray!20}95.32 \\

    \bottomrule
    \end{tabular}
    \caption{\df{Coverage and accuracy scores for the \texttt{She} and \texttt{He} classes in the en$\rightarrow$it direction. The upper three rows report results based solely on automatic string-matching evaluation (as in Table~\ref{tab:performance}), while the lower three rows also include OOC instances where gender translation was manually evaluated.}}
    \label{tab:revised_accuracy}
\end{table}

\section{Data Statistics}
\label{sec:data_statistics}

\begin{table*}[!t]
\footnotesize
\centering
\begin{tabular}{l|ccc|ccc|ccc}
\toprule
& \multicolumn{3}{c|}{\textbf{\# Samples}} & \multicolumn{3}{c|}{\textbf{\# Hours}} & \multicolumn{3}{c}{\textbf{\# Speakers}} \\
 & \cellcolor{gray!20}\textbf{All} & \textbf{\texttt{She}} & \textbf{\texttt{He}} & \cellcolor{gray!20}\textbf{All} & \textbf{\texttt{She}} & \textbf{\texttt{He}} & \cellcolor{gray!20}\textbf{All} & \textbf{\texttt{She}} & \textbf{\texttt{He}} \\
\midrule
\texttt{train} & \cellcolor{gray!20}5061 & 2514 & 2546 & \cellcolor{gray!20}9.70 & 5.08 & 4.62 & \cellcolor{gray!15}1417 & 534 & 883 \\
\texttt{dev} & \cellcolor{gray!20}1078 & 539 & 539 & \cellcolor{gray!20}2.08 & 1.10 & 0.98 & \cellcolor{gray!20}337 & 132 & 205 \\
\texttt{test-generic} & \cellcolor{gray!20}3702 & 1026 & 2676 & \cellcolor{gray!20}6.69 & 1.89 & 4.80 & \cellcolor{gray!20}38 & 11 & 27 \\
\texttt{test-speaker} (en$\rightarrow$es) & \cellcolor{gray!20}555 & 271 & 284 & \cellcolor{gray!20}1.33 & 0.69 & 0.64 & \cellcolor{gray!20}202 & 104 & 98 \\
\texttt{test-speaker} (en$\rightarrow$fr) & \cellcolor{gray!20}595 & 304 & 291 & \cellcolor{gray!20}1.36 & 0.71 & 0.65 & \cellcolor{gray!20}189 & 88 & 101 \\
\texttt{test-speaker} (en$\rightarrow$it) & \cellcolor{gray!20}551 & 271 & 280 & \cellcolor{gray!20}1.31 & 0.68 & 0.63 & \cellcolor{gray!20}147 & 69 & 78 \\

\bottomrule
\end{tabular}
\caption{Statistics of number of samples, hours, and speaker for the data splits used.}
\label{tab:data_stats}
\end{table*}

Table~\ref{tab:data_stats} summarizes the distribution of samples, hours, and speakers across the dataset splits used for training and testing our probes.

The \texttt{train} set consists of $5061$ samples, covering $9.70$ hours of audio, with a balanced gender representation across $1417$ speakers. However, the number of speakers in the \texttt{He} class is higher than in the \texttt{She} class ($883$ vs. $534$), reflecting the inherent imbalance in the MuST-C dataset \cite{mustc}.
The \texttt{dev} set, approximately $20\%$ the size of \texttt{train}, contains $1078$ samples and $2.08$ hours of gender-balanced audio from $337$ speakers.
The \texttt{test-generic} set includes $3702$ samples spanning $6.69$ hours, with a lower speaker-to-sample ratio than \texttt{train} and \texttt{dev}. Unlike these sets, \texttt{test-generic} is not gender-balanced, as it was created by merging the original test and validation sets of MuST-C, where male-speaker samples are significantly more numerous.
Finally, the \texttt{test-speaker} set is the smallest, containing over $550$ samples and $1.30$ hours of audio, with small variations across language directions. Unlike \texttt{test-generic}, this set maintains gender balance.

\begin{figure*}[t!]
    \centering
    \begin{subfigure}[b]{0.47\textwidth}
        \includegraphics[width=\textwidth]{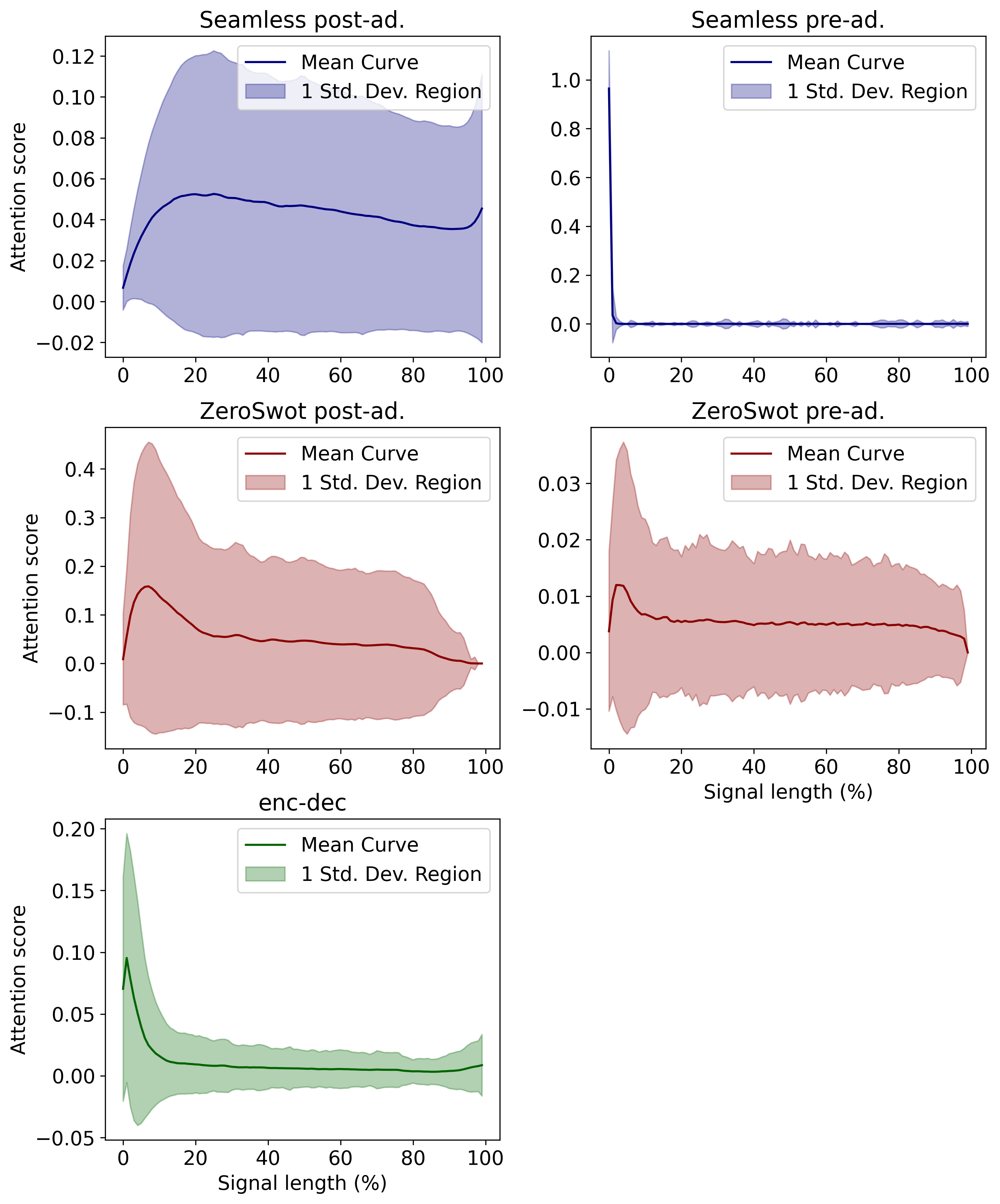}
        \subcaption{\texttt{test-generic}}
    \end{subfigure}
    \hfill
    \begin{subfigure}[b]{0.47\textwidth}
        \includegraphics[width=\textwidth]{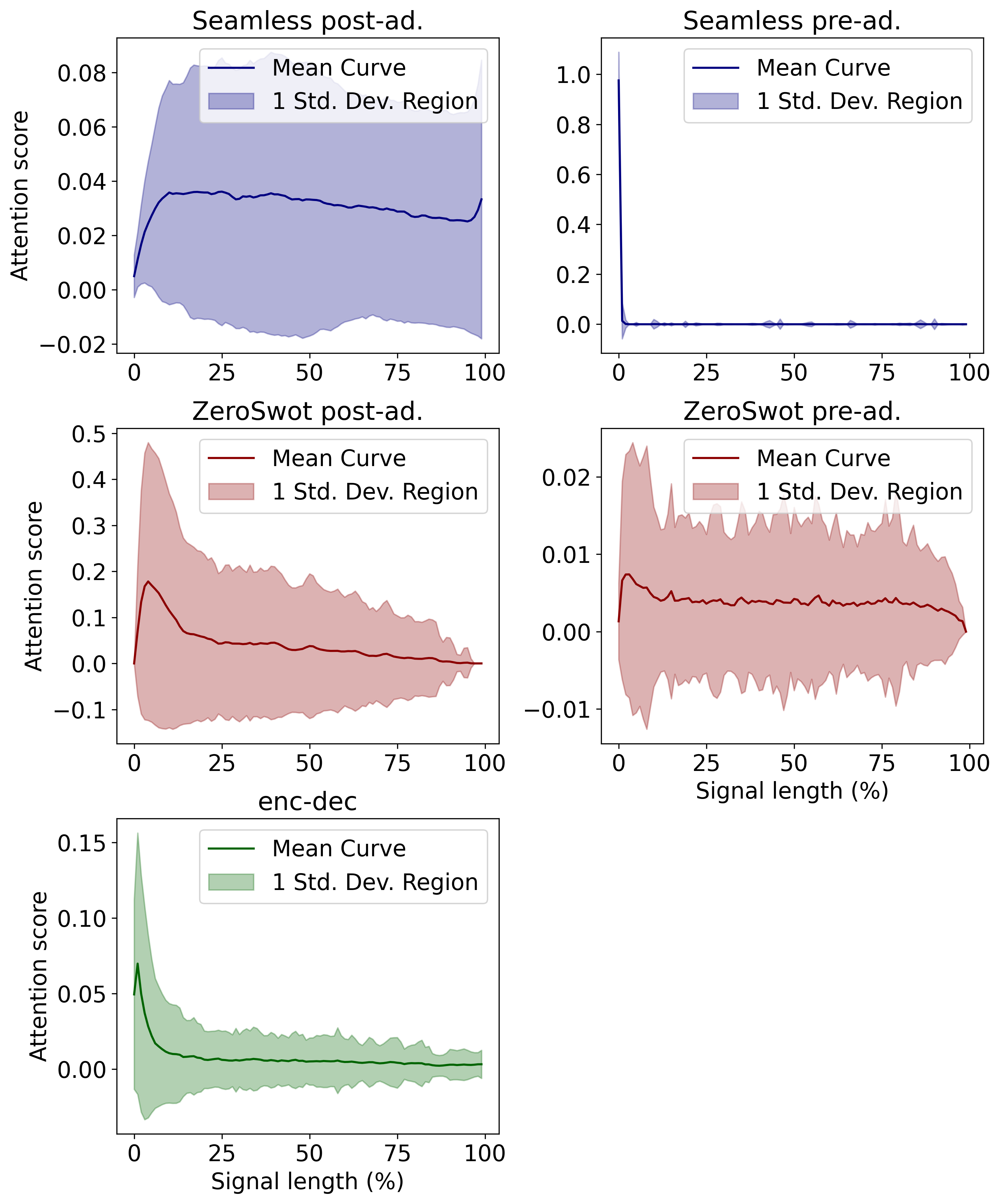}
        \subcaption{\texttt{test-speaker} (average across language pairs)}
    \end{subfigure}
    \caption{
    \df{Mean attention weights with standard deviations for the various attention-based probes.}
    }
    \label{fig:correlation_side_by_side}
\end{figure*}

\section{Gender Encoding Across Sequence Lengths}
\label{app:gender_in_seq_len}

For each input sequence, 
our probes provide attention weights \( \mathbf{a} \) that indicate the positions in the sequence where the probe primarily focuses when predicting gender (see \S \ref{sec:gender_probing}). 
These trends can serve as a proxy for understanding how gender information is distributed across the sequence length, particularly for probes that effectively extract gender-related features.
To analyze the overall weight distribution, we compute the average attention weight distribution over all test instances. However, since hidden state sequences vary in length depending on input audio duration, we interpolate the attention weight sequences to a fixed length of $100$ elements. Figure \ref{fig:correlation} shows the average distribution of these resampled attention weights for all considered probes.

In probes trained on the \texttt{enc-dec}, \texttt{Seamless pre-ad}, and \texttt{ZeroSwot pre-ad/post-ad} representations, we observe an initial peak in attention followed by a descending trend. However, \texttt{Seamless pre-ad} shows a sharp peak at the very beginning of the sequence, suggesting that only the earliest portion is primarily used for classification.
In contrast, for \texttt{Seamless post-ad}, where the probe performs worst, the initial positions appear to be the least relevant, although no clear pattern emerges. This could be due to the absence of gender encoding, leading to suboptimal probe performance.

Overall, our findings suggest that the earliest positions in the sequence are \df{particularly important for predicting gender, especially when the probe is highly effective.}
This aligns with previous work by \citet{krishnan24_interspeech} and \citet{attanasio-etal-2024-twists}, \df{who}
found that gender encoding is predominantly concentrated at the beginning of the sequence.
However, it is important to note that the specific positions attended to can vary depending on the probe, which may develop ineffective strategies, especially when performance is suboptimal.

\begin{figure*}[t!]
    \centering
    \begin{subfigure}[b]{0.28\textwidth}
        \centering
        \includegraphics[width=\textwidth]{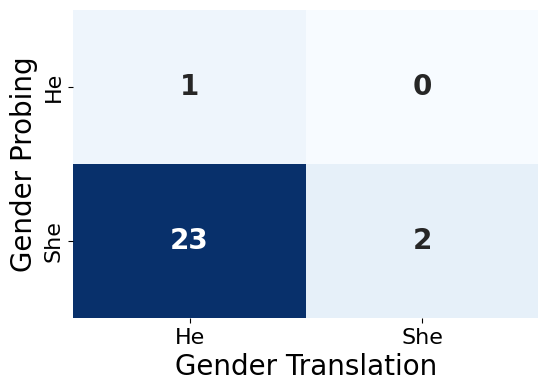}
        \caption{en-es}
    \end{subfigure}
    \hfill
    \begin{subfigure}[b]{0.28\textwidth}
        \centering
        \includegraphics[width=\textwidth]{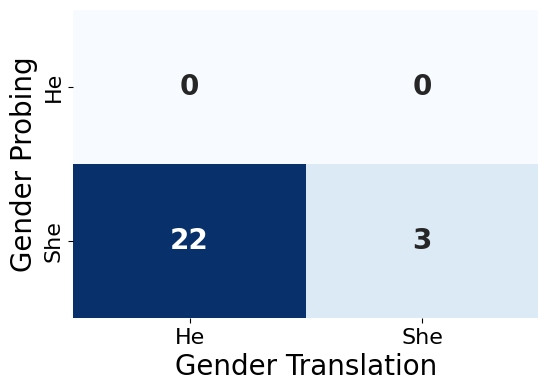}
        \caption{en-fr}
    \end{subfigure}
    \hfill
    \begin{subfigure}[b]{0.28\textwidth}
        \centering
        \includegraphics[width=\textwidth]{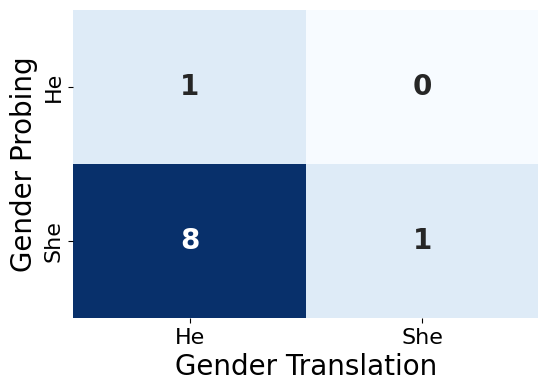}
        \caption{en-it}
    \end{subfigure}
    \caption{Confusion matrices showing the relationship between \textit{incorrect} probe predictions (gender probing) and corresponding gender translation.}
    \label{fig:conf_matrix_wrong}
\end{figure*}

\begin{figure*}[t!]
    \centering
    \begin{subfigure}[b]{0.28\textwidth}
        \centering
        \includegraphics[width=\textwidth]{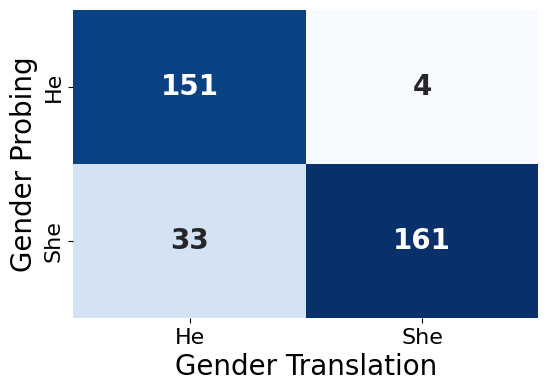}
        \caption{en-es}
    \end{subfigure}
    \hfill
    \begin{subfigure}[b]{0.28\textwidth}
        \centering
        \includegraphics[width=\textwidth]{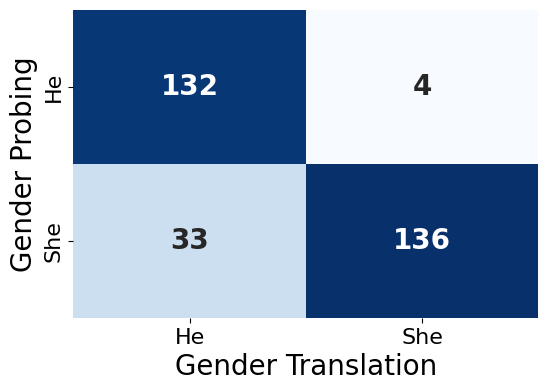}
        \caption{en-fr}
    \end{subfigure}
    \hfill
    \begin{subfigure}[b]{0.28\textwidth}
        \centering
        \includegraphics[width=\textwidth]{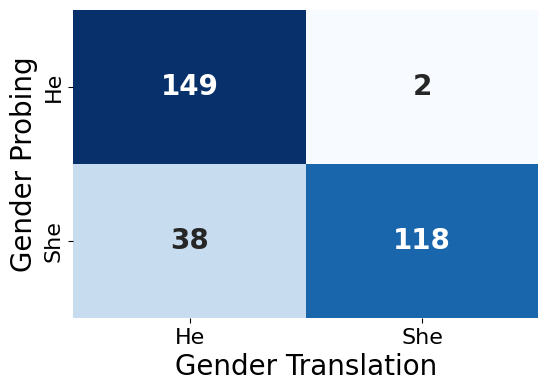}
        \caption{en-it}
    \end{subfigure}
    \caption{Confusion matrices showing the relationship between \textit{correct} probe predictions (gender probing) and corresponding gender translation.}
    \label{fig:conf_matrix_correct}
\end{figure*}

\begin{table*}[t!]
\centering
\footnotesize
\setlength{\tabcolsep}{3pt}
\begin{tabular}{l|llll}
    \toprule
    \multirow{3}{*}{\textit{a.}} & \multirow{3}{*}{en-es} & \multirow{3}{*}{\texttt{She}}  & \texttt{SRC} & They have formed me as a democratic citizen and a \textbf{bridge builder}. \\
    &  &  & \texttt{REF} & Me han formado como ciudadana democrática y \textbf{constructora de puentes}. \\
    &  &  & \texttt{OUT} & Me han formado como ciudadana democrática y como \textbf{un constructor de puente}. \\
    \midrule
    \multirow{3}{*}{\textit{b.}} & \multirow{3}{*}{en-es} & \multirow{3}{*}{\texttt{He}}   & \texttt{SRC} & And just as the woman who wanted to know me as \textbf{an adult} got to know me, she turned into a [...]. \\
     &  &   & \texttt{REF} & Y así como la mujer que quería conocerme como \textbf{adulto} llegó a conocerme, se convirtió en una [...]. \\
     &  &   & \texttt{OUT} & Y así como una mujer que quería conocerme como \textbf{adulta} se conoció a mí, se convirtió en una [...]. \\
     \midrule

    \multirow{3}{*}{\textit{c.}} & \multirow{3}{*}{en-fr} & \multirow{3}{*}{\texttt{He}}   & \texttt{SRC} & And I wound up getting involved with the space community, really \textbf{involved} with NASA, [...]. \\
    &  &   & \texttt{REF} & Et j'ai fini par être impliqué dans la communauté spatiale, réellement \textbf{impliqué} avec la NASA, [...]. \\
    & &   & \texttt{OUT} & Et j'ai fini par m'impliquer dans la communauté spatiale, vraiment \textbf{impliquée} par la NASA, [...]. \\
    \midrule
    \multirow{3}{*}{\textit{d.}} & \multirow{3}{*}{en-fr} & \multirow{3}{*}{\texttt{She}}  & \texttt{SRC} & My main sport was soccer, and I was \textbf{a goalkeeper}, which is [...]. \\
    &  &   & \texttt{REF} & Mon sport principal était le football et j'étais \textbf{gardienne de but}, ce qui est [...]. \\
    &  &   & \texttt{OUT} & Mon principal sport était le foot et j'étais \textbf{un gardien}, qui est [...]. \\
    \midrule

    \multirow{3}{*}{\textit{e.}} & \multirow{3}{*}{en-it} & \multirow{3}{*}{\texttt{She}}  & \texttt{SRC} & And I want to say something a little bit radical for \textbf{a feminist}, and that is that I think that [...]. \\
    &  &   & \texttt{REF} & Vorrei dire qualcosa di abbastanza radicale per \textbf{una femminista}, cioè che credo [...]. \\
    &  &   & \texttt{OUT} & E voglio dire qualcosa di un po' radicale per \textbf{un femminista}, e penso che [...]. \\
    \midrule
    \multirow{3}{*}{\textit{f.}} & \multirow{3}{*}{en-it} & \multirow{3}{*}{\texttt{She}}  & \texttt{SRC} & As a researcher, \textbf{a professor} and a new parent, my goal is to [...]. \\
     &  &  & \texttt{REF} & Da ricercatrice, \textbf{professoressa} e adesso genitore, il mio obiettivo è [...]. \\
    &  &  & \texttt{OUT} & Da ricercatrice, \textbf{professore} e genitore nuovo, il mio obiettivo è [...]. \\
\bottomrule
\end{tabular}
\caption{
\df{Transcribed source sentences (\texttt{SRC}) from \texttt{test-speaker}, accompanied by reference translations (\texttt{REF}) and output translations (\texttt{OUT}) generated by \texttt{enc-dec}. Misgendered words in \texttt{OUT}, along with their correct counterparts in \texttt{REF} and aligned terms in \texttt{SRC}, are shown in bold.}
}
\label{tab:examples}
\end{table*}

\section{A Closer Look at Divergencies Between Gender Probing and Translation}
\label{app:fine-grained}

In \S \ref{sec:results_gender_translation}, we show that gender encoding correlates with the model's gender translation ability. 
\df{For \texttt{enc-dec}, both gender encoding and gender translation achieve good accuracy, although the probe's accuracy is higher. 
As a result, some translations are incorrect even when gender can be accurately predicted from the hidden states.
}

\df{To better understand these discrepancies,}
we plot confusion matrices. 
\df{Figure \ref{fig:conf_matrix_wrong} shows the confusion matrix comparing \textit{incorrect} probe predictions with the corresponding translations by the \texttt{enc-dec} model, while Figure~\ref{fig:conf_matrix_correct} displays the confusion matrix for \textit{correct} probe predictions alongside their associated translations.}
\df{First, we observe that in Figure~\ref{fig:conf_matrix_wrong}, the number of instances is low, as the probes are generally effective at predicting gender for \texttt{enc-dec}. Among the mismatches, the most frequent pattern involves the probe incorrectly predicting feminine while the model correctly translates masculine---$23$ cases for en$\rightarrow$es, $22$ for en$\rightarrow$fr, and $8$ for en$\rightarrow$it.
In contrast, Figure~\ref{fig:conf_matrix_correct} shows that most discrepancies occur when the ST model translates as masculine despite the probe correctly predicting \texttt{She}: $33$ cases for both en$\rightarrow$es and en$\rightarrow$fr, and $38$ for en$\rightarrow$it. The reverse case---feminine translations for the \texttt{He} class---is much rarer, with only $4$ cases each for en$\rightarrow$es and en$\rightarrow$fr, and $2$ for en$\rightarrow$it.
This asymmetry reflects the persistent gender bias in translation, which affects even the \texttt{enc-dec} model despite its overall strong performance on gender translation.}

To investigate why the ST model produces incorrect translations despite correctly encoding gender, we manually examined the divergent cases from Figure \ref{fig:conf_matrix_correct}. We identified several patterns where the model struggles, and provide examples in Table \ref{tab:examples}.
\df{In certain cases, stereotypical associations---particularly involving professions---appear to influence the translations.}
For instance, in \textit{a)}, \textit{d)}, and \textit{f)} terms like ``bridge builder'', ``goalkeeper'', and ``professor''---often associated with men---are translated as masculine even when referring to women, despite other gendered words in the sentence being correctly translated as feminine.
In other cases, the model 
\df{appears to struggle}
with complex agreement resolutions. For example, in \textit{c)}, ``involved'' may have been linked to ``space community'' due to proximity, rather than the speaker. In \textit{e)}, ``for a feminist'' may not have been correctly interpreted as referring to the speaker. In \textit{b)}, the presence of another gendered referent (``woman'' and ``she'') may have led the model to misinterpret ``adult'' as feminine.

Overall, these findings suggest that linguistic biases can sometimes override acoustic cues, limiting the model’s ability to fully leverage gender information.
However, a more systematic analysis is needed in future work, which is beyond the scope of this paper.

\end{document}